\documentclass{article}

\usepackage{microtype}
\usepackage{graphicx}
\usepackage{subfigure}
\usepackage{subcaption}
\usepackage{caption}
\usepackage{enumitem}
\usepackage{sidecap}
\usepackage{multirow}
\usepackage{multicol}
\usepackage{arydshln}
\usepackage[table]{xcolor}
\usepackage{framed}
\usepackage{booktabs} 

\usepackage{hyperref}

\usepackage[accepted]{icml2025}

\usepackage{amsmath}
\usepackage{amssymb}
\usepackage{mathtools}
\usepackage{amsthm}

\usepackage[capitalize,noabbrev]{cleveref}

\theoremstyle{plain}
\newtheorem{theorem}{Theorem}[section]

\theoremstyle{definition}
\newtheorem{definition}[theorem]{Definition}

\theoremstyle{remark}

\theoremstyle{definition}
\newtheorem{optimization}{Optimization}

\usepackage[textsize=tiny]{todonotes}

\icmltitlerunning{MIXER: Mixed Hyperspherical Random Embedding Neural Network for Texture Recognition}

\newcommand{\specialparagraph}[1]{\textbf{#1}.~}
\newcommand{\Bold}[1]{\mathbf{#1}}

\begin{document}

\twocolumn[
\icmltitle{MIXER: Mixed Hyperspherical Random Embedding Neural Network for Texture Recognition}


\icmlsetsymbol{equal}{*}

\begin{icmlauthorlist}
\icmlauthor{Ricardo T. Fares}{yyy}
\icmlauthor{Lucas C. Ribas}{yyy}
\end{icmlauthorlist}

\icmlaffiliation{yyy}{Institute of Biosciences, Humanities and Exact Sciences, São Paulo State University, São José do Rio Preto, Brazil}

\icmlcorrespondingauthor{Lucas C. Ribas}{lucas.ribas@unesp.br}

\icmlkeywords{Representation Learning, Randomized Neural Networks, Feature Extraction}

\vskip 0.3in
]



\printAffiliationsAndNotice{}  

\begin{abstract}
Randomized neural networks for representation learning have consistently achieved prominent results in texture recognition tasks, effectively combining the advantages of both traditional techniques and learning-based approaches.
However, existing approaches have so far focused mainly on improving cross-information prediction, without introducing significant advancements to the overall randomized network architecture.
In this paper, we propose \textsc{Mixer}, a novel randomized neural network for texture representation learning. 
At its core, the method leverages hyperspherical random embeddings coupled with a dual-branch learning module to capture both intra- and inter-channel relationships, further enhanced by a newly formulated optimization problem for building rich texture representations. 
Experimental results have shown the interesting results of the proposed approach across several pure texture benchmarks, each with distinct characteristics and challenges.
The source code will be available upon publication.
\end{abstract}

\section{Introduction}

Texture is a widely studied visual feature in both research fields of computer vision \cite{haralick1973textural} and neuroscience \cite{julesz1981textons}. Its study has not only enabled a deeper understanding of human texture perception but also allowed computers to efficiently represent them in their own language for later use in \emph{downstream tasks}, such as material recognition \cite{sharan2013recognizing,chen2021deep}, land classification \cite{akiva2022self}, and biological image analysis \cite{oiticica2025using}.

To tackle these tasks, many texture representation approaches have been proposed in the literature, which currently can be broadly classified in three groups: \emph{handcrafted} \cite{haralick1973textural,manjunath1996texture,leung2001representing,ojala2002multiresolution}, \emph{learning-based} \cite{cimpoi2015deep,zhang2017deep,zhai2020deep}, and \emph{hybrid} techniques \cite{junior2016elm,scabini2023radam,su2023lightweight}. The first ones are characterized by approaches manually designed from scratch, which provide greater explainability but generally have limited descriptive power. On the other hand, the second group is characterized by the use of neural networks to automatically extract the essential attributes. Although they offer greater descriptive power, these methods are generally less interpretable and demand substantial amounts of data and energy. 

In this context, the third group of approaches seeks to leverage the advantages of the previous ones while simultaneously addressing their inherent limitations.
In this group, approaches are characterized by those that either combine traditional descriptors with neural networks as in \citet{su2023lightweight} or those as in \citet{junior2016elm,scabini2023radam} that use neural networks while maintaining inherent characteristics of some \emph{handcrafted} descriptors, such as the absence of a dataset to work properly as for instance in LBP \cite{ojala2002multiresolution}, i.e., given an image a LBP representation is promptly returned without the need of a pre-defined training and test split.
Therefore, thinking in a Venn diagram, \emph{hybrid} approaches can be seen as the intersection of the \emph{handcrafted} and \emph{learning-based} approaches, such that some of the mentioned \emph{learning-based} techniques may be considered \emph{hybrid} ones.

Currently, interesting hybrid texture recognition approaches are those based on shallow and fast neural networks. In particular, to the best of our knowledge, \citet{junior2016elm} proposed the first texture representation learning technique using a Randomized Neural Network (RNN). This is a shallow 1-hidden-layer neural network, where the hidden layer is randomly initialized and the output layer weights are computed using a closed-form solution. This leads to a rapid training procedure, and the resulting weights are subsequently used to compose the image representation.
This technique has been successfully applied in representation learning tasks, such as in color-texture \cite{ribas2024color}, and shape representation \cite{junior2018randomized,fujiwara2020neural,ribas2024learning}.

However, most of these representation learning techniques have focused primarily on constructing improved input and output feature matrices to enhance the learning of weights through better cross-information prediction, that is, guiding the network to learn more effective weights by forcing it to predict more complex relationships between input and output content. In contrast, little to no research in texture representation learning has simultaneously explored architectural improvements, such as enhancements to the random projector or the underlying optimization problem.

In this work, we introduce \textsc{Mixer}, a novel texture representation learning random neural network. 
It is composed of four core modules. For any given image, the first module, called the \emph{local pattern extractor}, is responsible for densely extracting small patches from each image channel independently, aiming to capture raw local texture information.
Next, the second module, called the \emph{hyperspherical random projector}, encodes the extracted patches into random embeddings and constrains them to lie on the surface of a hypersphere through embedding normalization. The third module, referred to as the \emph{learning} module, consists of two branches that capture intra- and inter-channel intensity relationships. The first branch employs a purely randomized autoencoder, while the second addresses this learning through a novel optimization problem.
Finally, the last module, called \emph{compression} module, compresses (summarizes) the decoders' learned weights into a useful color-texture representation for recognition tasks.

\specialparagraph{Contributions of this paper} Our contributions are summarized as follows:
\begin{itemize}[leftmargin=*]
    \item We introduced a novel hyperspherical multi-head random projector that: (i) distinctly encodes the local patches of different image channels, thereby implicitly introducing channel ownership information; and (ii) constrains the random embeddings to lie on the surface of a hypersphere, retaining only their directional information.
    
    \item We proposed a novel learning module composed of the \textsc{Direct} and \textsc{Mixed} learning branches. The former captures intra-channel intensity relationships, whereas the latter focuses on inter-channel relationships, achieved through a new optimization problem applied to an intermediate fused representation.

    \item Experimental results demonstrated that our color-texture representation consistently outperformed the existing approaches, being the only among the compared approaches to surpass the 97\% barrier of Outex, and the 99\% barrier of the average accuracy.
\end{itemize}

\specialparagraph{Structure of the paper} Section \ref{sect:background} presents the notation and the required background. Section \ref{sect:method} introduces the method of our novel texture representation. Section \ref{sect:experiments} exhibits the experimental setup. Section \ref{sect:results_and_discussions} shows the results and presents the discussion. To conclude, Section \ref{sect:conclusions_and_future_works} terminates the work with the conclusions and future works.

\section{Background and Notation} \label{sect:background}

\specialparagraph{Notation} Throughout the paper, we adopt the following notation conventions for readability. Let the lowercase letters $a \in \mathbb{R}$ be scalars, and the boldface lowercase letters $\Bold{a} \in \mathbb{R}^{n}$ be $n$-dimensional vectors, where $a_k \in \mathbb{R}$ is its $k$-th component. We define the boldface uppercase letters $\Bold{A} \in \mathbb{R}^{m \times n}$ as a real matrix of size $m \times n$, where $a_{ij} \in \mathbb{R}$ is its $(i, j)$-entry. Further, let calligraphic uppercase letters $\mathcal{A}$ denote a set of elements, with the following exceptions that $\mathbb{R}^n$ is the set of all $n$-dimensional real vectors, $\mathbb{R}^{m \times n}$ is the set of all real matrices of size $m \times n$, and $\mathbb{N}$ is the set of natural numbers.

\subsection{Texture Recognition}

\specialparagraph{Classical Texture Recognition} The early days of texture recognition relied heavily on \emph{classical} or \emph{handcrafted} feature extractors, which were manually designed by specialists and typically based on predefined steps. 
To begin with, the pioneering work by \citet{haralick1973textural} not only proposed a feature extraction technique based on measures such as angular second moment, contrast, and correlation derived from gray-level co-occurrence matrices, but also introduced the notion of texture as the spatial distribution of gray-level tones, although no formal and universally accepted definition exists to this day.

Later, studies on the visual information processing system \cite{daugman1985uncertainty,jones1987evaluation} discussed the receptive fields of simple cells in the visual cortex, showing that they can be expressed in terms of Gabor filters.
This inspired researchers in proposing texture feature extractors using these filters or wavelets \cite{bovik1990multichannel,manjunath1996texture}. 
In general, these approaches consist of generating a filter bank with filters of distinct orientations and frequencies, and then convolving the image with each filter.
Subsequently, measures such as the mean and standard deviation can be computed from the filter responses to serve as features of the input image.

Following this, \citet{ojala1996comparative} proposed the Local Binary Pattern (LBP). At its core, a local neighborhood scans the entire image with overlap, and the values within each neighborhood are thresholded relative to the central pixel value.
The numbers resulting from the binary patterns of each neighborhood are aggregated into a histogram, which serves as the texture representation.
In addition, many improved and robuster versions and similar techniques to LBP were proposed, such as \citet{ojala2002multiresolution} that introduced the widely known rotation invariant version of the former approach, \citet{ojansivu2008blur} proposed the Local Phase Quantization (LPQ) technique, a texture descriptor robust to blurring, and \citet{guo2010completed} presented the complete LBP (CLBP).

Around the early 2000s, techniques such as Bag of Textons (BoT) \cite{leung2001representing} and Bag of Visual Words (BoVW) \cite{csurka2004visual} emerged, representing an image as a histogram over a learned or predefined codebook \cite{liu2019bow}.
Generally, given a training set, these methods select and describe local patches from the images and then build a codebook by applying a vector quantization algorithm, such as $k$-means, over the set of local patch descriptors.
Later, this codebook is used to generate the histogram from a test image, which serves as its feature vector. In addition, with the rise of the deep-learning models, the phase of local patch description turned to use deep-learned features, such as in FV-CNN \cite{cimpoi2015deep}, DeepTEN \cite{zhang2017deep} and BoFF \cite{florindo2023boff}.

\specialparagraph{Deep-learning-based Texture Recognition} 
Since the rapid emergence of deep learning methods in the last decade, the research community has seen a variety of approaches adopting Convolutional Neural Networks (CNNs) and, more recently, Vision Transformers (ViTs) for texture recognition.
This is largely attributed to their ability to automatically learn and extract key features from images, in contrast to the manually designed descriptors discussed earlier.

These approaches can be divided into two categories: feature extraction and end-to-end learning.
The former is characterized by the use of the extracted features from a off-the-shelf pre-trained backbone followed by the use of a classic classifier, such as KNN and SVM. 
The latter involves training a complete architecture from scratch or fine-tuning a pre-trained model.

Remarkable examples of feature extraction approaches are: FV-CNN \cite{cimpoi2015deep} which uses the Fisher Vector (FV) to pool the local features from a convolutional layer of a CNN backbone, ignoring the spatial information, thus creating an orderless representation; and RADAM \cite{scabini2023radam} that randomly encodes multi-depth aggregated features from CNN backbones to later compose the texture representation. 

Conversely, in the end-to-end learning category, notable techniques are: DeepTEN \cite{zhang2017deep} where the authors proposed a single network that performs the feature extraction, the codebook (or dictionary) learning (in a supervised manner) and the encoding of the representation; DEPNet \cite{xue2018deep} where authors discussed that the textures in surfaces preserve some spatial information. 
Thus, guided by this starting point, the authors proposed the texture encoding branch and the global average pooling branch. The former addresses the orderless representation of the image, while the latter captures local spatial cues.
Later, DSRNet \cite{zhai2020deep} is a texture recognition network that learns the spatial dependency between texture primitives. This was motivated by the fact that multiple texture primitives appear differently in their spatial context; however, they present an inherent 
spatial dependency.
Following, \citet{chen2024enhancing} proposed the DTP (Deep Tracing Pattern) module. This module is responsible for leveraging simultaneously shallow and deep layer features, thus obtaining rich cross-layer features for texture recognition. To this end, the DTP module perform a cross-layer feature aggregation, and inspired by LBP, it densely samples 3D patches of the grouped tensor to capture spatial and cross-layer information. After, a spatial pyramid histogram is used to comprise the feature vector.

\subsection{Randomized Neural Network}

\specialparagraph{General Background} Randomized neural networks proposed in \cite{huang2006extreme,pao1992functional,pao1994learning,schmidt1992feed} are continuously being used by the research community for pattern recognition tasks. In particular, canonical application areas are \textit{classification} and \textit{regression}. However, as previously mentioned, in the last decade novel approaches employing randomized neural networks have been proposed for representation learning. These methods have been applied primarily to textures, dynamic textures, and shape representations, diverging from their more common applications.

In general, the large adoption of the community is due to three essential aspects that attracted interest in randomized neural networks. In the first place, the simplicity, since the most common cases are architectures composed by a single fully-connected hidden layer with randomly generated weights, and whose responsibility is to perform a non-linear random projection of the input data. 
Another important aspect is its speed and low computational cost, which result from computing the output layer weights through a closed-form solution. This enables fast training without the need for backpropagation. Finally, these models also exhibit good generalization capabilities for certain types of problems.

Commonly, there are two widely utilized randomized neural networks in the literature:  Randomized Vector Functional Link (RVFL), and the Extreme Learning Machine (ELM). Both architectures share the common aspect of non-linearly random project the input data into another dimensional space, and the characteristic of a fast-training phase conducted via a closed-form solution by using the (regularized) least-squares solution.

Nevertheless, their main difference lies in how the output nodes are connected. The RVFL architecture has the output nodes connected both with the hidden nodes and the input ones, while in the ELM architecture, the output nodes are only connected with the hidden ones. 
Although its strengths and weaknesses have been extensively discussed, particularly regarding how the output layer should be connected.
In this work, the output layer is connected only with the hidden nodes, following the previous literature in texture representation learning using randomized neural networks.

\specialparagraph{Formalism} In this part, we present a mathematical background of the randomized neural network.

Let $\mathcal{D} = \{ (\Bold{x}_i, \Bold{y}_i) \}_{i=1}^{N} \subset \mathcal{X} \times \mathcal{Y}$ be the dataset consisting of $N$ samples, where $\Bold{x}_i \in \mathcal{X}$ is a $p$-dimensional input feature vector, and $\Bold{y}_i \in \mathcal{Y}$ is a $r$-dimensional output feature vector.
From $\mathcal{D}$ it is assembled the input and output feature matrices represented by $\Bold{X} = \text{hconcat}(\Bold{x}_1, \Bold{x}_2, \dots, \Bold{x}_N) \in \mathbb{R}^{p \times N}$, and $\Bold{Y} = \text{hconcat}(\Bold{y}_1, \Bold{y}_2, \dots, \Bold{y}_N) \in \mathbb{R}^{r \times N}$, respectively, where $\text{hconcat}(\cdot)$ is the horizontal concatenation operation.

Following this, the network non-linearly projects these input feature vectors into another dimensional space.
To this end, a matrix $\boldsymbol{\psi} \in \mathbb{R}^{\omega \times p}$ is randomly generated by some strategy, such as sampling each matrix's entry from some distribution, e.g., $\boldsymbol{\psi}_{ij} \sim \mathcal{N}(\mu, \sigma^2)$, or using some pseudorandom number generator procedure. Further, defining this matrix is a fundamental step, because the number of neurons in the hidden layer ($\omega$), more specifically the dimension of the space where the features are projected into, shall be specified. In this context, upon the specification of the random weight matrix $\boldsymbol{\psi}$, the hidden or projected feature matrix is given by:
\begin{equation} \label{eq:proj_matrix}
    \Bold{Z} = \alpha(\boldsymbol{\psi}\Bold{X}) \in [0, 1]^{\omega \times N}\,,
\end{equation}
where $\alpha(x) := 1/(1 + e^{-x})$ is the sigmoid function applied element-wise. In general, one can choose another non-linear function, such as hyperbolic tangent.
Accordingly, from Equation \eqref{eq:proj_matrix}, the projected matrix consists of all real vectors $\Bold{z}_i = \alpha(\boldsymbol{\psi}\Bold{x}_i) \in \mathbb{R}^{\omega}$, which represent the $\omega$-dimensional internal representation of the input feature vector $\Bold{x}_i \in \mathbb{R}^{p}$.

Subsequently, the learning phase to obtain the trained weights of the output layer shall be conducted. In this sense, the main objective of a randomized neural network is to seek for a matrix $\boldsymbol{\varphi} \in \mathbb{R}^{r \times \omega}$ of a linear transformation $T : \mathbb{R}^{\omega} \to \mathbb{R}^{r}$, that takes the image of $\Bold{z}_i$, being $\Bold{z}_i$ the internal representation of $\Bold{x}_i$, as close as possible to $\Bold{y}_i$ under the squared $\ell_2$ norm.
More precisely, this involves solving the following minimization problem:
\colorlet{shadecolor}{gray!10}
\begin{shaded}
\begin{optimization}[Least-squares]
    Let $\Bold{X} \in \mathbb{R}^{p \times N}$, $\Bold{Y} \in \mathbb{R}^{r \times N}$ and $\Bold{Z} \in \mathbb{R}^{\omega \times N}$ be the previously defined input, output, and projected feature matrices. We seek for a matrix $\boldsymbol{\varphi}^* \in \mathbb{R}^{r \times \omega}$ that solves the following minimization problem:
    \begin{equation} \label{eq:least_squares}
        \underset{\boldsymbol{\varphi}}{\min} \; \lVert \Bold{Y} - \boldsymbol{\varphi} \Bold{Z} \rVert_F^2 + \gamma \, \Omega(\boldsymbol{\varphi}) \,,
    \end{equation}
    where $\lVert \cdot \rVert_F$ is the Frobenius norm, $\Omega(\cdot)$ is the regularization term, and $\gamma > 0$ is the regularization parameter.
\end{optimization}
\end{shaded}

In particular, when the regularization term is set to $\Bold{\Omega}(\boldsymbol{\varphi}) = 0$, i.e., no regularization is applied. The solution for the minimization problem in Equation \eqref{eq:least_squares} is given by the least-squares solution: $\boldsymbol{\varphi}^* = \Bold{Y}\Bold{Z}^\dagger$, where $\Bold{Z}^\dagger = \Bold{Z}^T(\Bold{Z}\Bold{Z}^T)^{-1}$ is the Moore-Penrose pseudoinverse \cite{generalized1955penrose}.

Alternatively, a regularization term may be introduced for two main reasons. First, the regularization contributes in balancing the model complexity by adjusting the model's bias-variance trade-off. Second, inverting the matrix $\mathbf{Z}\mathbf{Z}^T$ may be challenging when it is ill-conditioned. Under these circumstances, one may utilize the Tikhonov's regularization \cite{calvetti2000tikhonov} without losing the closed-form aspect. To this end, the regularization term is set to $\Omega(\boldsymbol{\varphi}) = \lVert \boldsymbol{\varphi} \rVert_F^2$. In this case,  the solution of Problem \eqref{eq:least_squares} is given by the regularized least-squares solution:
\begin{equation}
    \boldsymbol{\varphi}^* = \Bold{Y}\Bold{Z}^T(\Bold{Z}\Bold{Z}^T + \gamma \Bold{Id})^{-1}\,,
\end{equation}
where $\Bold{Id}$ is the identity matrix of size $\omega \times \omega$.

\section{Method} \label{sect:method}

\begin{figure*}
    \centering
    \includegraphics[width=0.98\linewidth]{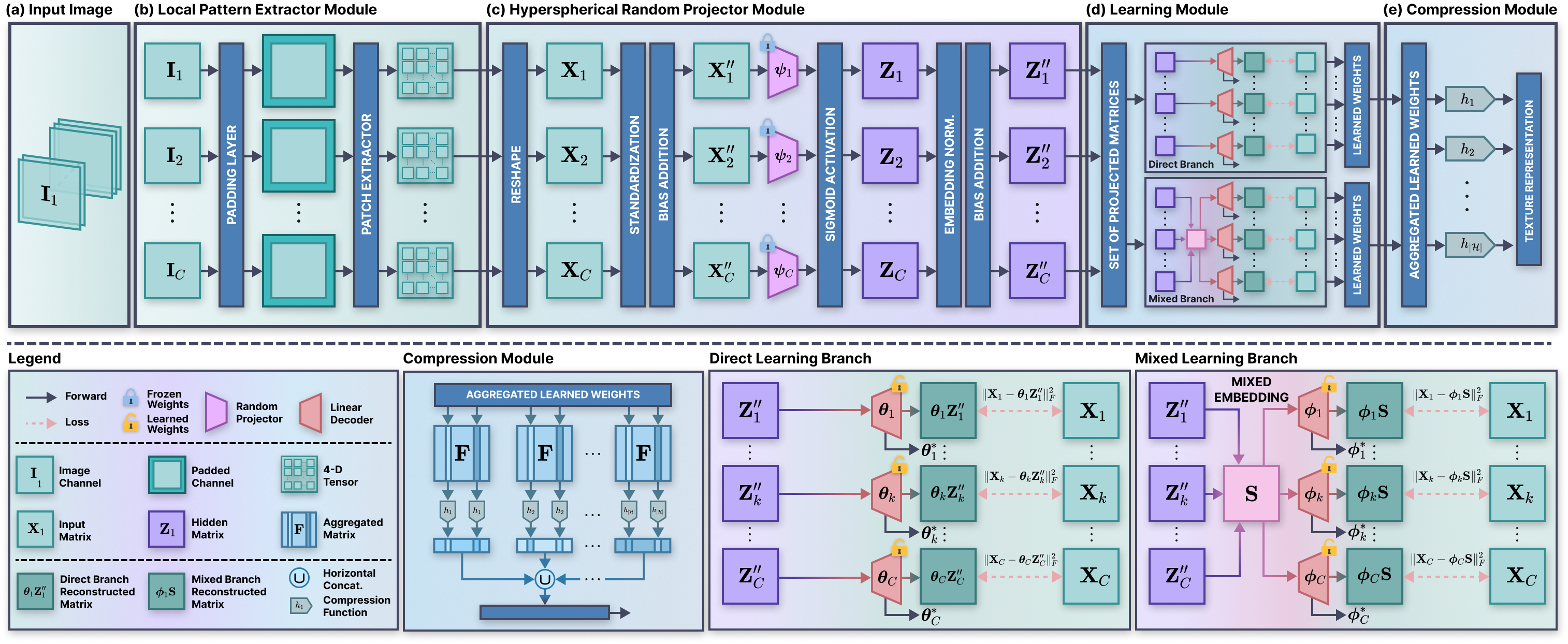}
    \caption{Overview of the \textsc{Mixer} pipeline. 
    The input image $\Bold{I} \in \mathbb{R}^{C \times H \times W}$ is fed to the Local Pattern Extractor (LPE) module, which pads the image and subsequently performs the extraction of tiny patches to record the raw texture information. 
    Thereafter, these patches are fed to the Hyperspherical Random Projector (HRP) module that maps these patches in hyperspherical random embeddings composing the random projected matrices $\Bold{Z}''_i \in \mathbb{R}^{(\omega + 1) \times HW}$.
    The projected matrices are fed to both \textsc{Direct} and \textsc{Mixed} branches responsible for learning the intra- and inter-channel local intensity relationships, respectively.
    The linear decoder's learned weights from both modules are fed to the compression module, which is responsible for vertically concatenating them, resulting in the aggregated learned weight matrix, and is responsible for applying selected compression functions to compress the weight matrix into a useful color-texture representation.
    }
    \label{fig:method}
\end{figure*}

In this section, we introduce \textsc{Mixer}, our novel approach for texture representation learning.
As illustrated in Figure \ref{fig:method}, \textsc{Mixer} consists of four core modules: 1) Local Pattern Extractor (LPE) module; 2) Hyperspherical Random Projector (HRP) module; 3) Learning module; and 4) Compression Module. 
Thus, the remainder of this section is dedicated to presenting the internal mechanisms of these modules.

\specialparagraph{Local Pattern Extractor Module} Given an image $\Bold{I} \in \mathbb{R}^{C \times H \times W}$. 
This module, depicted in Figure \ref{fig:method}(b), is responsible for extracting local intensity patterns from the image by densely sampling multiple small patches. Its purpose is to capture variations and implicit spatial relationships among pixel intensities, thereby inherently representing raw texture information to be subsequently used as input for the remainder of the network.

To this end, the module extracts multiple $J \times J$ patches centered at every pixel from each image channel of $\Bold{I}$, thus with overlapping.
However, centering a pixel at or near the image border is not feasible because the patch size would overflow the image bounds. 
Therefore, the image is padded before patch extraction.
Assuming that $J$ is an odd natural number, the padding size applied in all image channel sides is $\frac{J - 1}{2}$, and the utilized padding mode is replication. In general, this module is expressed by:
\begin{align}
     \label{eq:pad}
    & \text{Pad}_J(\Bold{I}) \in \mathbb{R}^{C \times (H + J - 1) \times (W + J - 1)} \\
    \label{eq:lpe}
    & \text{LPE}(\text{Pad}_J(\Bold{I})) \in \mathbb{R}^{C \times H \times W \times J \times J} \,.
\end{align}

Therefore, the output of Equation \eqref{eq:lpe} is a $5$-dimensional tensor that, basically, stores every $J \times J$ image patches centered at every pixel of the image channel of dimension $H \times W$ for every image channel among the $C$ available. 
Finally, this module generalizes the widely used patch-based extraction technique, extending it to images with an arbitrary number of channels, as commonly employed in RNN-based texture representation learning approaches \cite{junior2016elm,ribas2024learning}.
We refer the reader to Section \ref{supp:lpe} of the supplementary materials for further details about this module.

\specialparagraph{Hyperspherical Random Projector Module} This module, illustrated in Figure \ref{fig:method}(c), encodes the raw pixel intensities of the extracted patches into internal representations, referred to as hyperspherical random embeddings. These embeddings are subsequently used during the training phase of the proposed learning modules.
In this context, to obtain these internal representations, the module performs non-linear random projections of the patch intensities, encoding them into a generally higher $\omega$-dimensional space. These projections are then constrained to lie on the surface of a unit hypersphere through embedding normalization.
This normalization step helps computational and training stability \cite{zhang2023hyperspherical,wang2020understanding}, and here it does by adjusting potential column-norm distortions in the projected (hidden) matrix and by enhancing the conditioning of the matrix during the inversion step.

To accomplish this, let $\Bold{L} = \text{LPE}(\text{Pad}_J(\Bold{I}))$ be the $5$-dimensional tensor output of the previous layer, which contains the pixel intensities of the extracted patches for every channel of the image. We reshape this tensor by independently merging the image spatial dimensions and the patch size dimensions. Following this, we interchange these last two reshaped dimensions. We express these steps by:
\begin{equation}
    \Bold{X} = \text{Reshape}(\Bold{L}) \in \mathbb{R}^{C \times J^2 \times HW} \,.
\end{equation}
Hence, after the reshaping process, we obtain $\mathbf{X}_1, \mathbf{X}_2, \dots, \mathbf{X}_C$, where each $\mathbf{X}_k \in \mathbb{R}^{J^2 \times HW}$ is a matrix containing the flattened, densely extracted patches over the spatial dimensions of the $k$-th image channel.
Subsequently, each of these matrices is used during the random projection phase to obtain the internal representation of each flattened patch. To this end, we perform:
\begin{equation}
    \Bold{Z}_k = \alpha(\boldsymbol{\psi}_k\Bold{X}''_k) \in [0, 1]^{\omega \times HW}\,,
\end{equation}
where $\boldsymbol{\psi}_k \in \mathbb{R}^{\omega \in (J^2 + 1)}$ is the random weight matrix\footnote{We refer the reader to the Section \ref{supp:weight_gen} of the Supplementary Material for the random weight matrix generation process.} used only to project the $k$-th pre-processed input feature matrix:
\begin{equation}
    \Bold{X}''_k = \text{vconcat}(\{ -\Bold{1}_{HW}, \Bold{X}_k' \}) \in \mathbb{R}^{(J^2 + 1) \times HW} \,,
\end{equation}
where $\text{vconcat}(\cdot)$ is the vertical concatenation operation, $\Bold{1}_{HW} \in \mathbb{R}^{1 \times HW}$ is a row matrix with all entries set to one, representing the bias terms, and $\Bold{X}_k'$ is the unit normal scaled input feature matrix defined by:
\begin{equation}
    [\Bold{X}'_k]_{ij} = \dfrac{[\Bold{X}_k]_{ij} - \overline{[\Bold{X}_k]}_{i,:}}{s_{k,i} + \epsilon} \,,  \quad k \in \{1, 2, \dots, C\}\,,
\end{equation}
where $\textstyle \overline{[\Bold{X}_k]}_{i,:} = \frac{1}{HW}\sum_{j=1}^{HW}[\Bold{X}_k]_{ij}$ is the sample mean of the $i$-th row of the $k$-th input feature matrix $\Bold{X}_k$, $s^2_{k,i} = \textstyle \frac{1}{HW - 1}\sum_{j=1}^{HW}([\Bold{X}_k]_{ij} - \overline{[\Bold{X}_k]}_{i,:})^2$ is the sample variance, and $\epsilon = 10^{-10}$ is a small positive number to prevent division by zero. 

In this context, we have that $\Bold{Z}_k$ consists of all the $\omega$-dimensional random embedded patch intensities. Following this, before entering the learning modules, these projected feature matrices $\Bold{Z}_1, \Bold{Z}_2, \dots, \Bold{Z}_C$ have their column vectors projected onto the unit hypersphere $\mathbb{S}^{\omega - 1} = \{\Bold{v} \in \mathbb{R}^{\omega} \; | \; \lVert \Bold{v} \rVert_2 = 1\}$. This process is expressed by:
\begin{equation}
    [\Bold{Z}'_k]_{ij} = \frac{[\Bold{Z}_k]_{ij}}{\max (\lVert [\Bold{Z}_k]_{:,j} \rVert_2, \epsilon)}\,, \quad k \in \{1, 2, \dots, C\}\,,
\end{equation}
where $\lVert \cdot \rVert_2$ is the Euclidean norm, and $[\Bold{Z}_k]_{:,j}$ is the $j$-th column of $\Bold{Z}_k$.
Finally, as done for the input feature matrix, we also add a bias weight by vertically concatenating a row matrix filled with $-1$, such that the output of this module is:
\begin{equation}
    \Bold{Z}_k'' = \text{vconcat}(\{-\Bold{1}_{HW}, \Bold{Z}_k'\}) \in \mathbb{R}^{(\omega + 1) \times HW} \,.
\end{equation}
To simplify the notation, we denote $\Bold{Z}_k''$, the output of this module for a input feature matrix $\Bold{X}_k$, as $f_{\boldsymbol{\psi}_k}(\Bold{X}_k)$, where $f_{\boldsymbol{\psi}_k}(\cdot)$ is designated as the $k$-th random encoder whose random weight matrix is given by $\boldsymbol{\psi}_k \in \mathbb{R}^{\omega \times (J^2 + 1)}$.

\specialparagraph{Learning Module} This module represents one of the main components of our novel approach, and is shown in Figure \ref{fig:method}(d). This is partitioned into two learning branches called the \textsc{Direct} and \textsc{Mixed} branches. These are responsible for learning multiple linear decoder networks, where their learned layer weights (scalars) are used later by the compression (summarization) module to assemble a useful color-texture representation.
\begin{itemize}[leftmargin=*]
    \item \textsc{\textbf{Direct}}: This branch is responsible for learning a mapping that reconstructs the local intensity patterns of the extracted $J \times J$ patches from the hyperspherical random embeddings, with the intent to induce the network to learn the relationship between the hidden (projected) space and the intensities (input) space, from the point of view of the raw spatial texture present in the input matrix $\Bold{X}$.

    To this end, this branch learns the mapping through the following optimization problem.
    
    \colorlet{shadecolor}{gray!10}
    \begin{shaded}
    \begin{optimization}[Direct Branch]
        Let $g_{\boldsymbol{\theta}}(\Bold{Z}) = \boldsymbol{\theta}\Bold{Z}$ be a linear decoder network with parameters $\boldsymbol{\theta}$. We seek to learn the parameters $\boldsymbol{\theta}^*_k \in \mathbb{R}^{J^2 \times (\omega + 1)}$ for each image channel by solving the following minimization problem:
        \begin{equation} \label{eq:direct_opt}
            \underset{\boldsymbol{\theta}_k}{\min} \; \lVert \Bold{X}_k - g_{\boldsymbol{\theta}_k}(f_{\boldsymbol{\psi}_k}(\Bold{X}_k)) \rVert_F^2 + \gamma_D \lVert \boldsymbol{\theta}_k \rVert_F^2 \,,
        \end{equation}
        where $\gamma_D > 0$ is the regularization parameter of the direct branch. This optimization problem acts as a pure randomized autoencoder, where we force the network to learn how to reconstruct the original patches intensities from a non-linear random projection of themselves.
    \end{optimization}
    \end{shaded}
    The optimization problem presented in Equation \eqref{eq:direct_opt} has a closed-form solution given by the following regularized least-squares solution:
    \begin{equation}
        \boldsymbol{\theta}^*_k = \Bold{X}_k\Bold{Z}{''_k}^T(\Bold{Z}''_k\Bold{Z}{''_k}^T + \gamma_D \Bold{Id})^{-1} \,.
    \end{equation}
    Finally, the subscript is emphasized because a linear decoder network is found for each channel of the input image. Thus, a set of linear decoders that learned the spatial texture relationships of each channel independently is obtained. In this sense, this branch outputs the set $\mathcal{S}_D = \{\boldsymbol{\theta}^*_1, \boldsymbol{\theta}^*_2, \dots, \boldsymbol{\theta}^*_C\}$ that characterizes the local intensity relationships present in the input image channels.

    \item \textsc{\textbf{Mixed}}: Unlike the \textsc{Direct} branch that uses its own patch internal representation to reconstruct itself. This branch aims to reconstruct each patch not only from its own random embedding but also from the random embeddings of the corresponding patch across all image channels.

    In this context, this branch first constructs a shared representation by combining the internal representations, serving as an intermediate fusion phase. 
    Following this, the result of the mixing process is used as input to the decoder network, whose main role is to reconstruct the original patch intensities. 
    Therefore, the network is guided to reconstruct the original patch information of a specific image channel by leveraging all available information from the corresponding patch across all channels, thereby learning inter-channel relationships.
    
    To achieve this, this branch learns the reconstruction through the following optimization problem:
    \colorlet{shadecolor}{gray!10}
    \begin{shaded}
    \begin{optimization}[Mixed Branch]
        Let $g_{\boldsymbol{\phi}}(\Bold{Z}) = \boldsymbol{\phi}\Bold{Z}$ be a linear decoder network with parameters $\boldsymbol{\phi}$. We seek to learn the parameters $\boldsymbol{\phi}^*_k \in \mathbb{R}^{J^2 \times (\omega + 1)}$ for each image channel by solving the following minimization problem:
        \begin{equation} \label{eq:mixed_opt}
            \underset{\boldsymbol{\phi}_k}{\min} \; \lVert \Bold{X}_k - g_{\boldsymbol{\phi}_k}(\Bold{S}) \rVert_F^2 + \gamma_M \lVert \boldsymbol{\phi}_k \rVert_F^2 \,,
        \end{equation}
        where $\gamma_M > 0$ is the regularization parameter of the mixed branch, and $\Bold{S} \in \mathbb{R}^{(\omega + 1) \times N}$ is the mixed random embedding built by averaging all internal random embeddings obtained for each image channel. That is,
        \begin{equation} \label{eq:mixed_rep}
            \Bold{S} = \frac{1}{C}\sum_{j=1}^{C}f_{\boldsymbol{\psi}_j}(\Bold{X}_j) \,.
        \end{equation}
    \end{optimization}
    \end{shaded}
    Consequently, from Equation \eqref{eq:mixed_rep}, we have that the shared internal representation $\Bold{s}_k$ of some patch $k$, emerge as the average of the internal representations of the same patch for all image channels. Next, similarly to the preceding optimization problem, the closed-form solution for the Equation \eqref{eq:mixed_opt} is given by the following regularized least-squares solution:
    \begin{equation}
        \boldsymbol{\phi}^*_k = \Bold{X}_k\Bold{S}^T(\Bold{S}\Bold{S}^T + \gamma_M \Bold{Id})^{-1}\,.
    \end{equation}
    Finally, as in the direct branch, the output of this branch is a set of linear decoders $\mathcal{S}_M = \{\boldsymbol{\phi}^*_1, \boldsymbol{\phi}^*_2, \dots, \boldsymbol{\phi}^*_C\}$ trained to reconstruct the local intensity patterns of each image channel from a common mixed representation.
\end{itemize}

\specialparagraph{Compression Module} Given an image $\Bold{I} \in \mathbb{R}^{C \times H \times W}$ its image representation should be defined by a $n$-dimensional real vector that carries useful information about $\Bold{I}$ to be used in downstream tasks. Nevertheless, as was observed, a set $\mathcal{S} = \mathcal{S}_D \cup \mathcal{S}_M \subset \mathbb{R}^{J^2 \times (\omega + 1)}$ containing the learned linear decoders' weights is obtained as output of the learning module (i.e., previous module). In this context, the role of this module, illustrated in Figure \ref{fig:method}(e), is to compress (summarize) the learned weights of these linear decoders, represented by real matrices, in real vectors to serve as a useful color-texture representation. To this end, we begin by defining a compression function:
\colorlet{shadecolor}{gray!10}
\begin{shaded}
\begin{definition}[Compression Function]
Every scalar-valued function $h : A \subset \mathbb{R}^{n} \to \mathbb{R}$ that maps a $n$-dimensional real vector $\Bold{x} \in A$ to a scalar $h(\Bold{x}) \in \mathbb{R}$, where $h(\Bold{x})$ embeds some structural or semantic information about $\Bold{x}$ is called a compression (summarization) function.
\end{definition}
\end{shaded}

By defining the compression function, also called the summarization function, we express that $h$ must extract some useful knowledge (simple or complex) about the input vector. 
In this sense, a function that receives any real vector and simply returns a random number cannot be considered a compression function, since it does not embed any information.

In this context, let $\Bold{F} = \text{vconcat}(\mathcal{S}) \in \mathbb{R}^{|\mathcal{S}|J^2 \times (\omega + 1)}$ be the real matrix resulting from the vertical concatenation of every matrix in $\mathcal{S}$, where $|\mathcal{S}| = 2C$, since $C$ decoders are learned in each branch. As a result, the matrix $\Bold{F}$ contains the weights of the all decoders learned in both proposed branches. Since these weights contain valuable content about the texture information of the input image, a set of summarization functions $\mathcal{H}$ are applied to compress the matrix $\Bold{F}$ in a real vector serving as representation of the texture image $\Bold{I}$. Therefore, this compression process is represented by:
\begin{align}
    \Bold{s}_h & = \left ( h(\Bold{f}_1), h(\Bold{f}_2), \dots, h(\Bold{f}_{\omega + 1}) \right ) \in \mathbb{R}^{\omega + 1} \\ 
    \label{eq:rep}
    \Bold{\Omega}_{\omega}(\Bold{I}) & = \text{hconcat}(\{ \Bold{s}_h \; | \; h \in \mathcal{H}\}) \in \mathbb{R}^{|\mathcal{H}|(\omega + 1)}\,,
\end{align}
where $\Bold{f}_k \in \mathbb{R}^{|\mathcal{S}|J^2}$ is the $k$-th column of $\Bold{F}$, $\Bold{s}_h$ is the real vector representing the compressed matrix $\Bold{F}$ under $h \in \mathcal{H}$, and $\Bold{\Omega}_\omega(\Bold{I})$ is the texture representation of the image $\Bold{I}$ using $\omega$-dimensional random embeddings.

In the present work, the set of compression functions employed is composed by four statistical measures $h_\mu$, $h_\sigma$, $h_\gamma$ and $h_\kappa$ which corresponds to the mean, standard deviation, skewness, and excess kurtosis, respectively. Thus, the dimensionality of the texture representation in Equation \eqref{eq:rep} is given by $4(\omega + 1)$. For more information about the employed statistical measures, such as their formulas, we refer the reader to Section \ref{supp:comp} of the supplementary material.

Finally, we assemble a new texture representation by late fusing the learned representations from distinct $\omega_1, \omega_2, \dots, \omega_n$ random embeddings sizes. This is a common utilized approach in previous RNN-based techniques, such as \cite{junior2016elm,ribas2020fusion}, which shown to improve the performance of the texture recognition task. Thus, we define this late fused color-texture representation as:
\begin{equation}
    \Bold{\Upsilon}_\mathcal{W}(\Bold{I}) = \text{hconcat}(\{ \Bold{\Omega}_\omega(\Bold{I}) \; | \; \omega \in \mathcal{W} \}) \in \mathbb{R}^{m}\,,
\end{equation}
where $\mathcal{W} = \{\omega_1, \omega_2, \dots, \omega_n\} \subset \mathbb{N}$ is the set containing the random embeddings sizes used to compose the late fused representation, and $\textstyle m = 4 \cdot \sum_{\omega \in \mathcal{W}}(\omega + 1)$ is the dimensionality of the proposed color-texture representation.

\section{Experiments} \label{sect:experiments}

In this section, we present the server infrastructure used to conduct the experiments, along with the versions of the main libraries employed to develop the proposed approach.
We then describe the benchmark datasets used to evaluate the technique, highlighting their main characteristics. Finally, we detail the cross-validation strategy, the evaluation metrics, and the classifier.

\specialparagraph{Setup} The experiments were conducted on a server running Ubuntu 22.04.03 LTS operating system equipped with a Intel i9-14900K processor, 128 GB of RAM, and a single graphic card NVIDIA RTX 4090. Specifically, the approach runs on the specified graphic card. Furthermore, the proposed technique is implemented using \textsc{PyTorch} v2.6.0, and the evaluation process used the \textsc{scikit-learn} v1.6.1 package.

\specialparagraph{Benchmarks} We evaluated our approach using four well-known texture datasets exhibiting distinct textural challenges, attributes, and characteristics. In this context, this enabled us to analyze the novel approach over a broad range of scenarios and how well the proposed approach generalized in the texture recognition task. Next, we briefly detail the peculiarities of each employed texture dataset:

\begin{itemize}[leftmargin=*]
    \item \textbf{OutexTC13} \cite{ojala2002outex}: This is the thirteenth texture classification suite proposed from the Outex surfaces and is widely used in both grayscale and colored versions. This dataset has proven to be hard to categorize due to the need for an efficient micro-textural characterization. Thus, as done in \cite{backes2013texture}, the employed dataset configuration is composed by 1360 color-texture images obtained by cropping 20 non-overlapping windows of size $128 \times 128$ pixels from 68 color-texture source images of size $538 \times 746$ pixels.

    \item \textbf{CUReT} \cite{dana1999reflectance}: It stands for Columbia-Utrecht Reflectance and Texture dataset. This dataset was proposed to study the appearances of the textured surfaces under distinct illuminations and viewing angles, therefore, tackling conditions that most datasets would not. In this sense, the presence of real-world surface textures (e.g., aluminum, leather, corduroy, paper etc.) under multiple light conditions and viewing angles makes the texture recognition task harder, thereby resulting in a suitable dataset for evaluation. The utilized dataset setup consists of 5612 color-texture images resulting from 61 existing classes with each 92 samples with dimensions $200 \times 200$ pixels. This is the same setup performed in \cite{scabini2020spatio}.
    
    \item \textbf{USPtex} \cite{backes2012color}: 
    This natural image dataset, built by the University of São Paulo focused on obtaining samples found day-to-day (e.g., rice, vegetation, walls, bricks etc.). The intent was to obtain a dataset with a higher degree of texture samples that exhibit irregular, random, or non-periodic patterns. The utilized configuration is composed by 2292 color-texture images divided in 191 classes, each containing 12 samples of dimensions $128 \times 128$ pixels.

    \item \textbf{MBT} \cite{abdelmounaime2013new}: The Multi-band Texture (MBT) dataset was proposed for the analysis of textural approaches focusing in samples with intra- and inter-band spatial variations. The colors present in these samples of MBT are due to the distinct textural content that comprises each color channel. In this context, this is an interesting dataset to evaluate how well our novel approach simultaneously describes these intra- and inter-band spatial variations. As done in \cite{scabini2020spatio}, we employed the dataset configuration that comprises 2464 samples with dimension $160 \times 160$ pixels by cropping 16 non-overlapping windows from each of the 154 original images of size $640 \times 640$ pixels.
\end{itemize}

\begin{figure}[t]
    \setlength{\tabcolsep}{-1pt}

    \centering
    \begin{tabular}{cc}
        \includegraphics[width=0.5\linewidth]{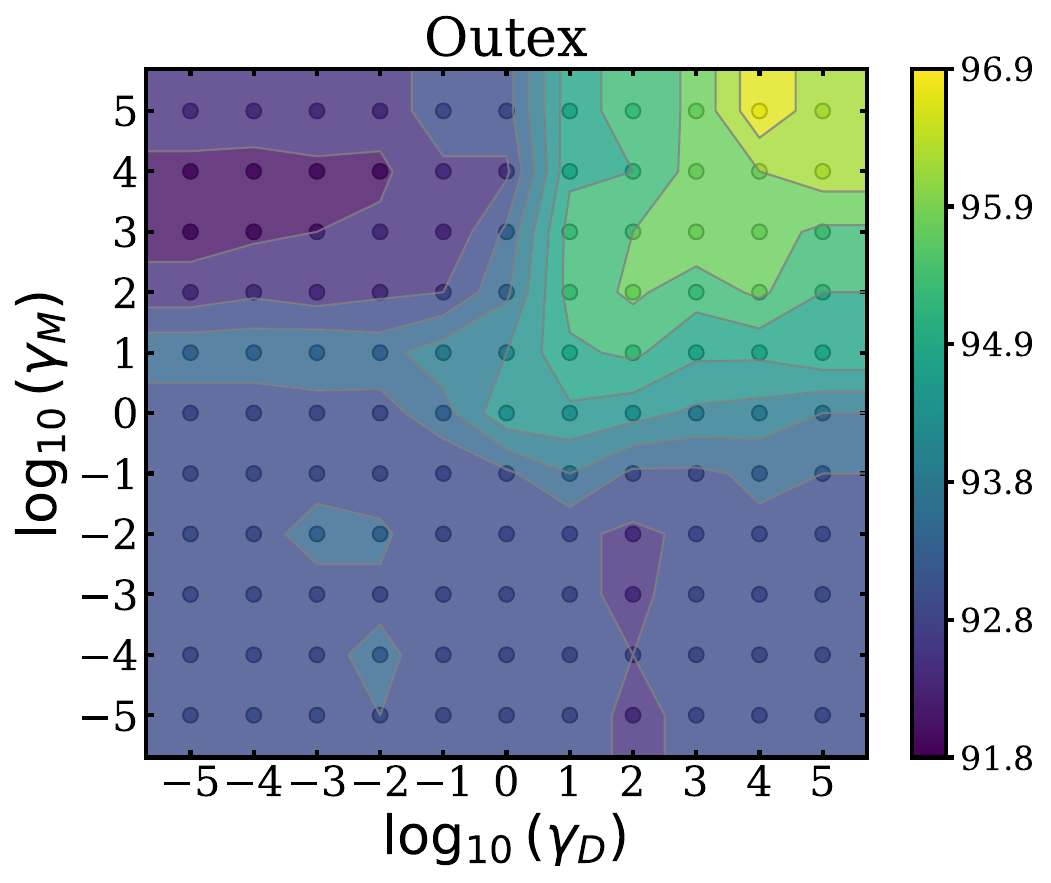} &
        \includegraphics[width=0.5\linewidth]{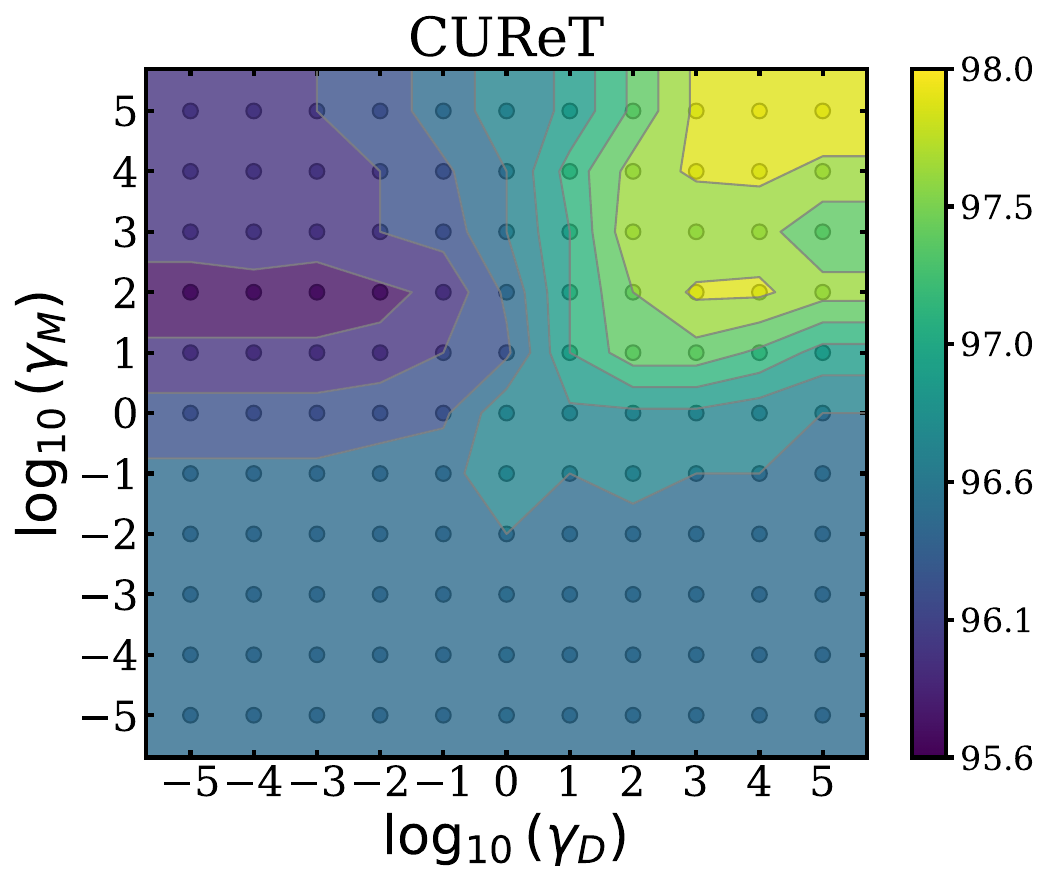} \\
        \includegraphics[width=0.5\linewidth]{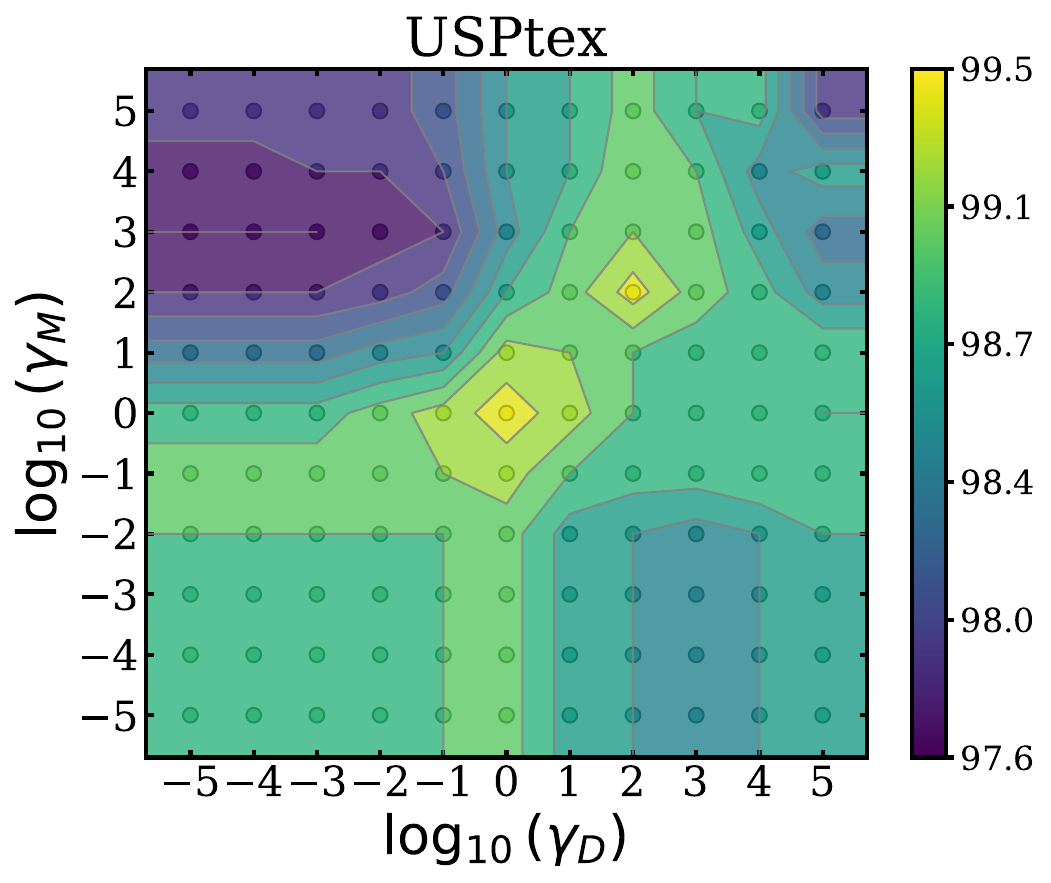} &
        \includegraphics[width=0.5\linewidth]{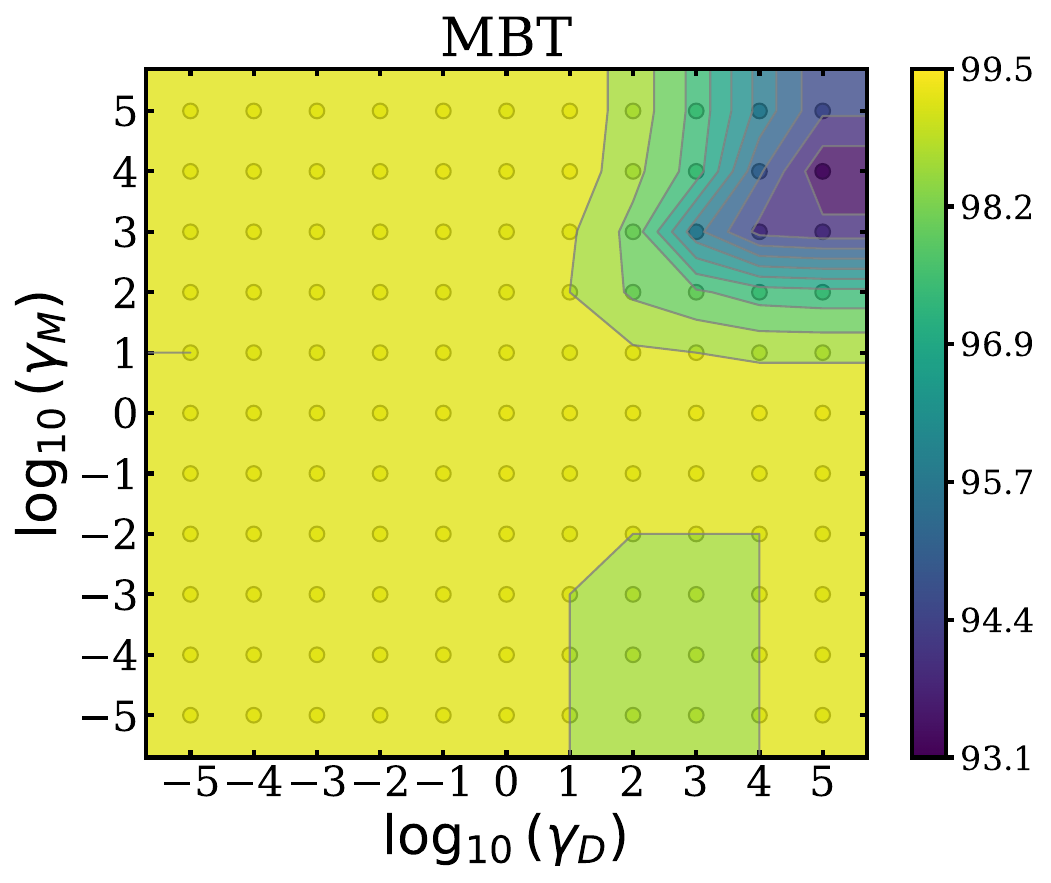}
    \end{tabular}
    
    \caption{Accuracy (\%) behavior of the color-texture representation $\Bold{\Omega}_{59}(\Bold{I})$ for all the benchmark datasets as the regularization of the \textsc{Direct} and \textsc{Mixed} branches varies.}
    \label{fig:regularization_analysis}
\end{figure}

\begin{figure*}[t]
    \setlength{\tabcolsep}{1.0pt}

    \centering
    \begin{tabular}{cc}
        \includegraphics[width=0.5\linewidth]{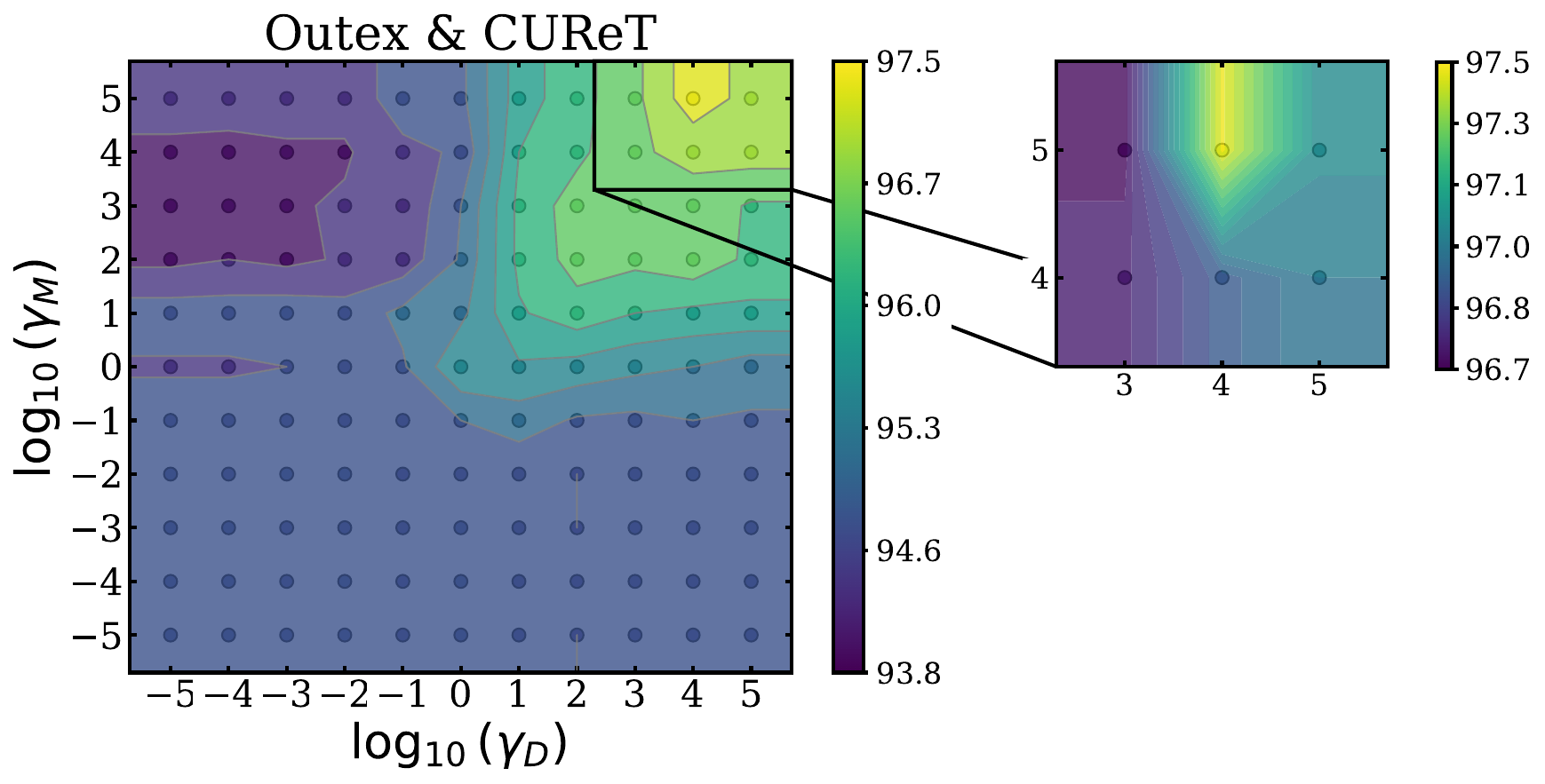} &
        \includegraphics[width=0.5\linewidth]{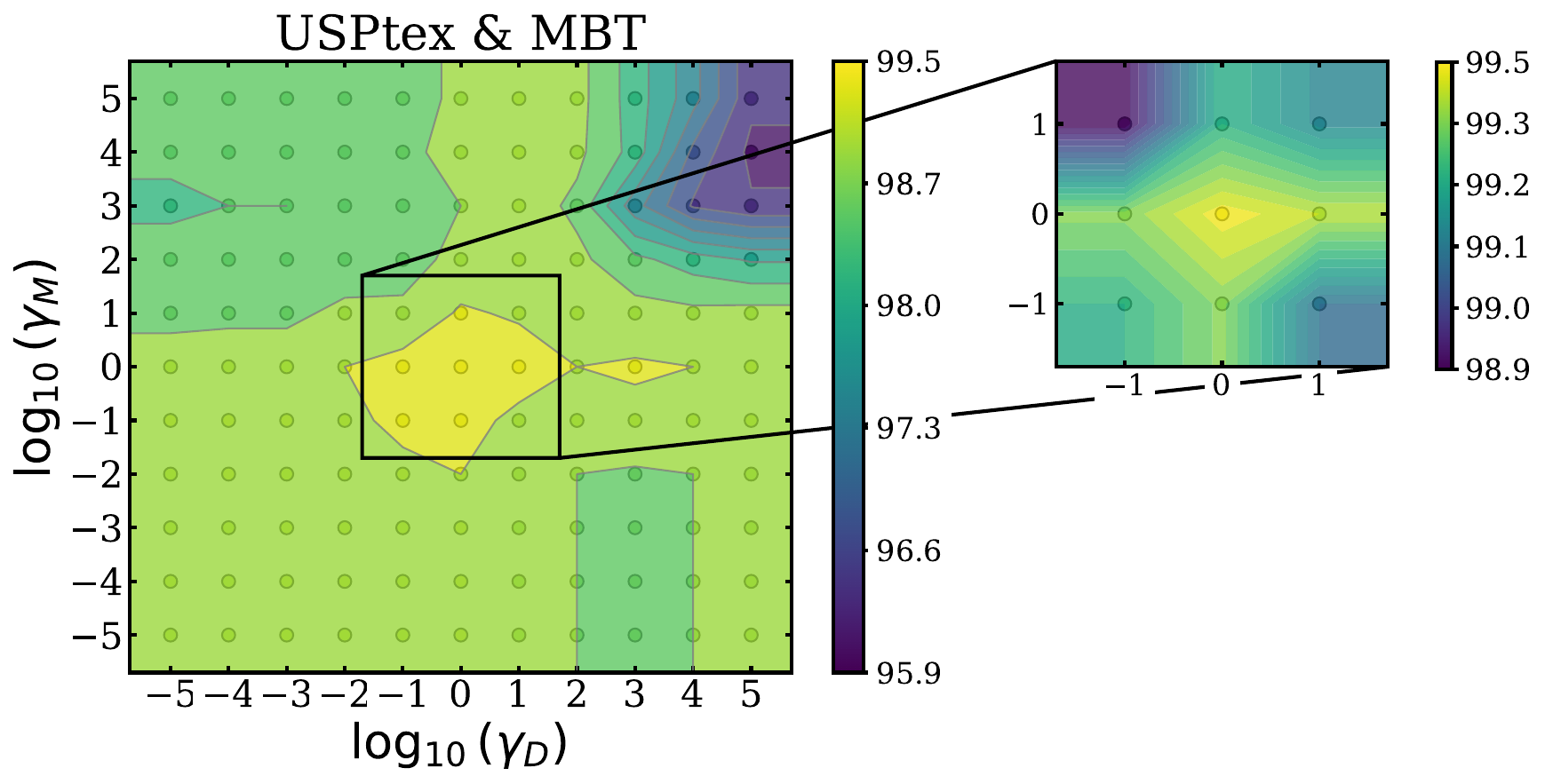}
    \end{tabular}

    \caption{Accuracy (\%) behavior of the color-texture representation $\Bold{\Omega}_{59}(\Bold{I})$ for the Outex \& CUReT and USPtex \& MBT dataset pairs as the regularization of the \textsc{Direct} and \textsc{Mixed} branches varies. The behavior is the average of the accuracies obtained by the datasets in each configuration. The inset plot in each figure refers to the region near to the highest average accuracy. 
    \textbf{Note}: The color bar of the inset plot was adjusted to help the visualization of the highest average accuracy.}
    
    \label{fig:regularization_analysis_combined}
\end{figure*}

\specialparagraph{Cross-validation, Metrics and Classifier}
To evaluate the performance of the novel approach we utilized the Linear Discriminant Analysis (LDA) classifier with the Leave-One-Out (LOO) cross-validation strategy using the accuracy as performance metric. This metric was used since all utilized benchmark datasets are balanced, and is a well-known performance proxy used for evaluation of texture recognition methods. In addition, we adopted this linear decision boundary classifier owing to its simplicity, and we used the default hyperparameter configuration provided by the \textsc{scikit-learn} v1.6.1 package, which uses the Singular Value Decomposition (SVD) as solver. For more information about the setting, we refer the reader to the supplementary material.


\section{Results and Discussions} \label{sect:results_and_discussions}

\specialparagraph{Regularization Analysis} To understand the effect of the regularization on the decoder's learned weights, and consequently, in the color-texture representation. We analyzed the behavior of the obtained accuracies of the texture representation $\Bold{\Omega}_{59}(\Bold{I})$ across all benchmark datasets, as the regularization hyperparameters $\gamma_D$ and $\gamma_M$ for the direct and mixed branches, respectively, varies over the region $\mathcal{R} \times \mathcal{R}$, where $\mathcal{R} = \{10^k \; | \; k \in \mathbb{Z} \; \land -5 \leq k \leq 5\}$.

Figure \ref{fig:regularization_analysis} exhibits a greater performance of the proposed color-texture representation in Outex and CUReT as the regularization values for both branches increased.
This is observed by warmer colors, representing greater accuracy, present on the upper right region, which is determined by high regularization on both branches. In that region, accuracies of up to 96.9\% were achieved in Outex, and up to 98.0\% in CUReT, thus indicating a good performance of the texture representation in this condition. In addition, another common aspect in these datasets was the relatively low performance identified by the presence of cooler colors in the remaining region, corresponding to the cases in which either one of the branches exhibited low regularization or both branches did.

These findings suggest that Outex and CUReT favor from the use of both branches, and while these are operating in conjunction, they had a stronger preference towards higher regularization to obtain good performances. This may be attributed to the fact that stronger regularization induces the decoders to reduce the fitting of the original patches intensities as well as possible noise, thus forcing the decoder to focus on the general discriminative information about the texture on the image, instead of the small peculiarities potentially present in it.

Following, USPtex shown a common aspect with the last two evaluated benchmarks, wherein all three presented a low performance in the upper left region, corresponding to low regularization on the \textsc{Direct} branch and high on the \textsc{Mixed} one. However, the fact that the proposed representation in USPtex shown a moderate performance when both regularization values were low, and this performance weakened as only the \textsc{Mixed} branch regularization increased indicates that this branch played a key role in the generalizability of the USPtex benchmark, and might be attributed to the fact that the increased regularization resulted in underfit, and consequently the decoder did not properly learn the textural content of the image under the mixing optimization problem, thus weakening the overall representation. In addition, unlike Outex and CUReT, USPtex has achieved the highest accuracy of 99.5\% with moderate values for regularization in both branches.

Subsequently, the performance behavior for the MBT dataset was basically the opposite of that observed in Outex and CUReT. The higher accuracies occurred outside the high regularization region, whereas the lower accuracies were obtained there. This denotes that both branches may be suffering from underfit when strongly regularized, as in the case of USPtex for the \textsc{Mixed} branch. Nevertheless, unlike the last three benchmarks, the MBT exhibited good performance on a broad range of regularization values, particularly those characterized by low to moderate levels. Among these regularization levels, the representation achieved the maximum accuracy of 99.5\%.

In addition to the preceding analysis, Figure \ref{fig:regularization_analysis} shows two pairs of datasets that exhibit similar suitable regularization values.
The first pair comprises Outex and CUReT, while the second consists of USPtex and MBT. Thus, to ensure fair hyperparameter selection within each pair, we adopt the same regularization values across the datasets.
We analyzed their average accuracies through a joint evaluation as the regularization values varied within the same parameter space.

Figure \ref{fig:regularization_analysis_combined} presents the joint analysis for both pairs of datasets.
For Outex and CUReT, the inset plot highlighting the region with the highest accuracy clearly shows that $\gamma_D = 10^4$ and $\gamma_M = 10^5$ yield the best average configuration.
Similarly, for USPtex and MBT, the plot reveals that $\gamma_D = 10^0$, and $\gamma_M = 10^0$ are the optimal configurations on average as well. Therefore, these configurations will be used for each pair for the remainder of the paper.

\begin{figure}[!htbp]
    \setlength{\tabcolsep}{1.0pt}

    \includegraphics[width=0.98\linewidth]{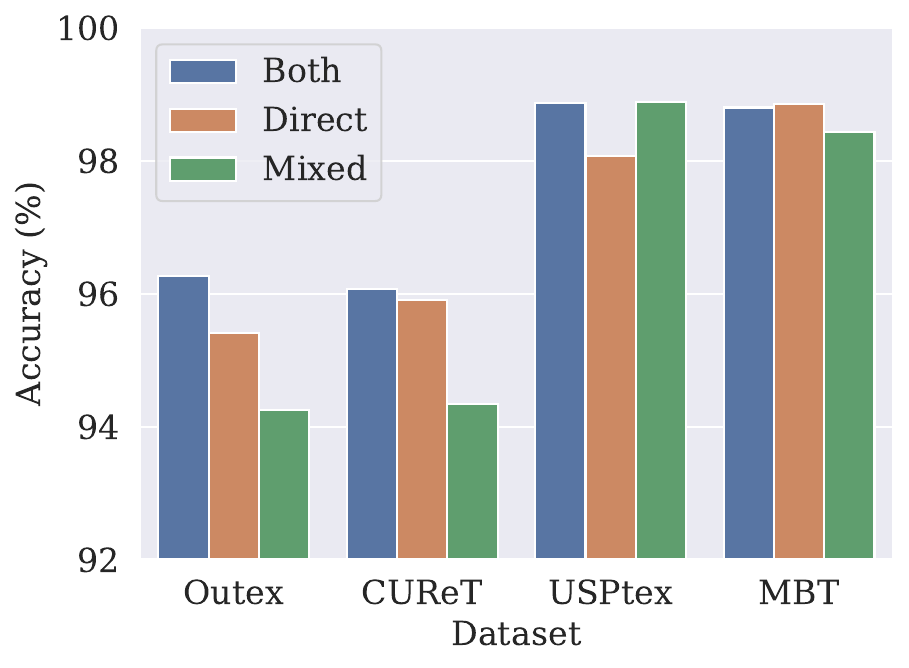}

    \caption{Average accuracy (\%) behavior of the proposed texture representation $\Bold{\Omega}_{\omega}(\Bold{I})$ when either one of the branches are being used or both of them are.}
    
    \label{fig:ablation_analysis}
\end{figure}

\specialparagraph{Ablation Analysis} Figure \ref{fig:ablation_analysis} presents the results for the ablation analysis of the proposed texture representation $\Bold{\Omega}_{\omega}(\Bold{I})$. In Outex and CUReT, it can be observed that the representation obtained using both branches increased the performance, being more notable for Outex. Furthermore, although the \textsc{Mixed} branch achieved the lowest accuracies on these datasets, the final representations were strengthened when both learning branches were combined. In other words, accuracy improved when both branches were used, suggesting that the features learned in each branch complemented one another.

In USPtex and MBT, it was observed that using only a single branch was sufficient to achieve the highest performance.
This indicates that one of the learning branches is already capable of extracting the meaningful features from the texture image. Interestingly, while USPtex shows the \textsc{Mixed} branch achieving the highest accuracy, MBT exhibits the best performance with the \textsc{Direct} branch.

Therefore, this brief analysis demonstrates the beneficial impact of employing both learning branches, particularly in the Outex and CUReT datasets, as well as the usefulness of having both branches available. This design highlights the adaptability of the approach to extract features from diverse textural patterns, as illustrated by the USPtex and MBT datasets, where a single branch was sufficient. However, the remainder of the paper uses the color-texture representation obtained when both learning branches are activated.

\specialparagraph{Embedding Size Analysis} We evaluated how the accuracy of the proposed texture representation $\Bold{\Omega}_{\omega}(\Bold{I})$ behaves as its unique parameter $\omega \in \mathbb{N}$, representing the random embedding size, varies. To this end, the parameter space where $\omega$ varied over was defined as $\mathcal{O} = \{ 9, 19, \dots, 109\}$, and is broader than those presented in previous studies \cite{junior2016elm,ribas2020fusion}. We also presented the obtained accuracies for all benchmark datasets, thereby providing a thorough analysis of the parameter impact.

\begin{SCfigure*}[][t]
    \centering
    \includegraphics[width=0.8\textwidth]{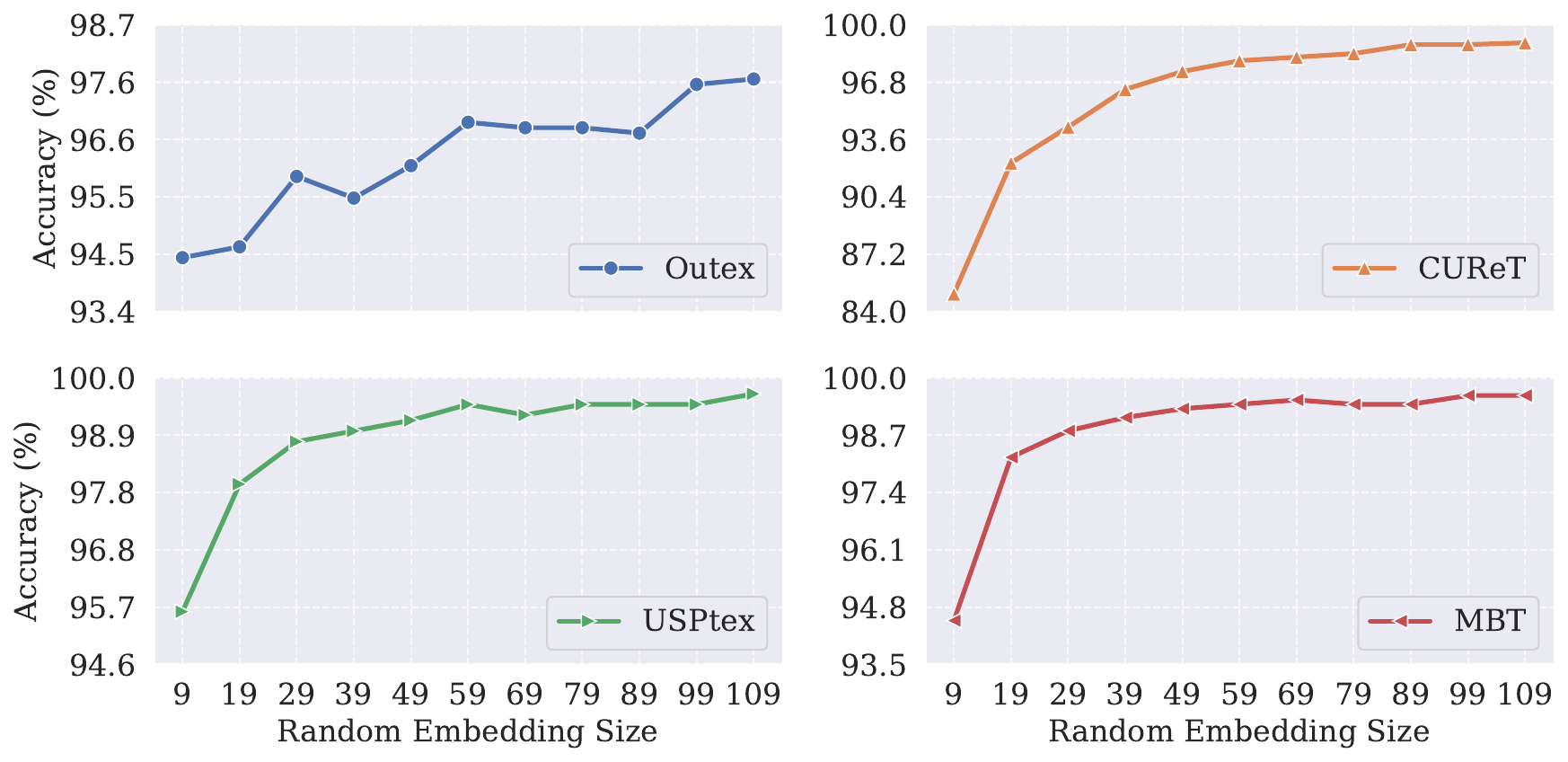}
    \caption{Accuracy (\%) behavior of the proposed texture representation $\Bold{\Omega}_{\omega}(\Bold{I})$ as the random embedding size $\omega$ varies in the defined parameter space. This behavior analysis is presented for all benchmark datasets.}
    \label{fig:parameter_analysis}
\end{SCfigure*}

Figure \ref{fig:parameter_analysis} presents the performance for all datasets. 
In particular, Outex generally exhibited an increase in accuracy as the embedding size grew. However, performance drops were observed in the intervals $[29, 39]\mathbb{N}$ and $[59, 89]\mathbb{N}$.
The former showed a more pronounced drop, while the latter exhibited only a subtle decrease. Conversely, CUReT, USPtex, and MBT, unlike Outex, displayed a sharp increase in performance at the initial values, followed by relative stability across the remaining parameter space.

This suggests that the embedding size had a beneficial impact on the overall performance of the proposed representation. This is largely motivated by the increasing number of learned features which depends on $\omega$, and which are discriminative enough to almost saturate the CUReT, USPtex and MBT simultaneously, and surpass the 97\% barrier, and almost the 98\% one, performance of the Outex dataset, where to the best of our knowledge is the unique RNN-based approach to achieve this as shall be discussed later.

In addition, following previous studies \cite{junior2016elm,ribas2020fusion}, we also investigated the performance of the novel approach $\Bold{\Upsilon}_{\mathcal{W}}(\Bold{I})$ when more random embedding sizes are incorporated.
We performed this analysis because previous studies have shown that combining features learned from lower-dimensional embeddings generally yields representations with higher performance, even when the total number of features remains the same.
 In particular, our study considered only two random embedding sizes to avoid generating excessively large feature vectors, with the selected pair of embedding sizes satisfying:

\begin{equation}
    (\omega_1, \omega_2) \in \{(\omega_1, \omega_2) \in \mathcal{O} \times \mathcal{O} \; | \; \omega_1 < \omega_2\} \,.
\end{equation}

Table \ref{table:combined_parameter_analysis} shows the performance of the approach for all benchmark datasets over the combined embeddings. Although there were a performance improvement of $\Bold{\Upsilon}_{\mathcal{W}}(\Bold{I})$ over $\Bold{\Omega}_{\omega}(\Bold{I})$, the increase was modest. Particularly, the greater improvement occurred in CUReT with 99.7\% $(\Bold{\Upsilon}_{99,109}(\Bold{I}))$ \textit{vs.} 99.0\% $(\Bold{\Omega}_{109}(\Bold{I}))$, followed by Outex 97.9\% $(\Bold{\Upsilon}_{39,79}(\Bold{I}))$ \textit{vs.} 97.7\% $(\Bold{\Omega}_{109}(\Bold{I}))$, and MBT 99.8\% $(\Bold{\Upsilon}_{69,79}(\Bold{I}))$ \textit{vs.} 99.6\% $(\Bold{\Omega}_{109}(\Bold{I}))$. There was no performance gain in USPtex. This suggests that our simpler approach $\Bold{\Omega}_{\omega}(\Bold{I})$ with only one embedding already contains enough discriminative features for classification.

Nevertheless, our experiments are consistent with previous findings indicating that combining lower-dimensional embeddings can yield better performance while maintaining the same number of features.
For instance, in Outex, the representation $\Bold{\Upsilon}_{29, 59}(\Bold{I})$ achieved 97.4\% with 360 descriptors, where the representation $\Bold{\Omega}_{89}(\Bold{I})$ achieved 96.7\% with the same amount of descriptors.

We end this section by selecting the texture representation that achieved the highest average accuracy, for a fairness comparison with the literature methods. This representation was $\Bold{\Upsilon}_{39,109}(\Bold{I})$ and achieved the average accuracy of 99.2\% with 600 descriptors.

\specialparagraph{Literature Comparison} We compared our novel approach with two types of texture recognition methods: the \textit{handcrafted} ones, and those based on \textit{randomized neural networks}. Included among the handcrafted are the Opponent-Gabor \cite{jain1998multiscale}, composed by opponent features from Gabor-filter outputs; the ones based in local patterns, such as LPQ \cite{ojansivu2008blur} and CLBP \cite{guo2010completed}; and those that models the texture as complex networks, building the representation upon network's topological measures, such as MCND \cite{scabini2019multilayer}, and SSN \cite{scabini2020spatio}.

%
%
\begin{table*}[!htb]
    \begin{minipage}{.5\linewidth}
        \centering

        \begin{tabular}{r c c c c c}
            \toprule
                                   & \multicolumn{5}{c}{Accuracy ($\uparrow$)} \\ \cline{2-6}
            $(\omega_1, \omega_2)$ & Outex13 & CUReT & USPtex & MBT & Avg.       \\ \hline
            
            \multicolumn{6}{c}{\textit{Line left for adjustment}} \\
            (9,19)   & 95.4 & 95.4 & 98.8 & 98.8 & 97.1 \\
            (9,29)   & 96.2 & 96.2 & 99.2 & 98.9 & 97.6 \\
            (9,39)   & 95.5 & 97.4 & 99.2 & 99.4 & 97.9 \\
            (9,49)   & 96.5 & 98.0 & 99.3 & 99.2 & 98.3 \\
            (9,59)   & 96.8 & 98.4 & 99.5 & 99.5 & 98.6 \\
            (9,69)   & 96.5 & 98.6 & 99.5 & 99.6 & 98.6 \\
            (9,79)   & 96.8 & 98.7 & 99.4 & 99.5 & 98.6 \\
            (9,89)   & 96.9 & 99.1 & 99.6 & 99.4 & 98.8 \\
            (9,99)   & 97.4 & 99.1 & 99.5 & 99.4 & 98.9 \\
            (9,109)  & 97.5 & 99.3 & 99.7 & 99.6 & 99.0 \\
            (19,29)  & 96.4 & 97.6 & 99.3 & 99.2 & 98.1 \\
            (19,39)  & 96.4 & 98.2 & 99.2 & 99.4 & 98.3 \\
            (19,49)  & 96.5 & 98.3 & 99.3 & 99.3 & 98.4 \\
            (19,59)  & 97.1 & 98.6 & 99.6 & 99.5 & 98.7 \\
            (19,69)  & 97.2 & 98.8 & 99.5 & 99.4 & 98.7 \\
            (19,79)  & 97.3 & 99.0 & 99.6 & 99.5 & 98.9 \\
            (19,89)  & 97.1 & 99.2 & 99.6 & 99.5 & 98.9 \\
            (19,99)  & 97.4 & 99.3 & 99.5 & 99.6 & 99.0 \\
            (19,109) & 97.5 & 99.2 & 99.7 & 99.7 & 99.0 \\
            (29,39)  & 96.6 & 98.4 & 99.6 & 99.4 & 98.5 \\
            (29,49)  & 96.8 & 98.5 & 99.5 & 99.3 & 98.5 \\
            (29,59)  & 97.4 & 98.9 & 99.6 & 99.3 & 98.8 \\
            (29,69)  & 97.3 & 98.9 & 99.6 & 99.6 & 98.9 \\
            (29,79)  & 97.6 & 99.1 & 99.5 & 99.5 & 98.9 \\
            (29,89)  & 97.3 & 99.3 & 99.7 & 99.4 & 98.9 \\
            (29,99)  & 97.1 & 99.3 & 99.7 & 99.4 & 98.9 \\
            (29,109) & 97.8 & 99.4 & 99.6 & 99.5 & 99.1 \\
            \bottomrule
        \end{tabular}

    \end{minipage}%
    \begin{minipage}{.5\linewidth}
        \centering

        \begin{tabular}{r c c c c c}
            \toprule
                                   & \multicolumn{5}{c}{Accuracy ($\uparrow$)} \\ \cline{2-6}
            $(\omega_1, \omega_2)$ & Outex13 & CUReT & USPtex & MBT & Avg.       \\ \hline
            
            (39,49)  & 97.1 & 98.8 & 99.6 & 99.5 & 98.8 \\
            (39,59)  & 97.6 & 99.1 & 99.6 & 99.7 & 99.0 \\
            (39,69)  & 97.6 & 99.1 & 99.6 & 99.7 & 99.0 \\
            (39,79)  & 97.9 & 99.2 & 99.5 & 99.6 & 99.1 \\
            (39,89)  & 97.1 & 99.3 & 99.7 & 99.5 & 98.9 \\
            (39,99)  & 97.4 & 99.4 & 99.5 & 99.6 & 99.0 \\
            (39,109) & 97.8 & 99.5 & 99.7 & 99.7 & 99.2 \\
            (49,59)  & 97.4 & 99.1 & 99.7 & 99.5 & 98.9 \\
            (49,69)  & 97.1 & 99.1 & 99.6 & 99.6 & 98.9 \\
            (49,79)  & 97.5 & 99.2 & 99.5 & 99.6 & 99.0 \\
            (49,89)  & 97.1 & 99.3 & 99.5 & 99.4 & 98.8 \\
            (49,99)  & 97.6 & 99.4 & 99.7 & 99.6 & 99.1 \\
            (49,109) & 97.6 & 99.4 & 99.5 & 99.6 & 99.0 \\
            (59,69)  & 97.6 & 99.2 & 99.7 & 99.6 & 99.0 \\
            (59,79)  & 97.6 & 99.5 & 99.6 & 99.6 & 99.1 \\
            (59,89)  & 97.4 & 99.5 & 99.6 & 99.6 & 99.0 \\
            (59,99)  & 97.6 & 99.5 & 99.5 & 99.6 & 99.1 \\
            (59,109) & 97.4 & 99.6 & 99.6 & 99.7 & 99.1 \\
            (69,79)  & 97.8 & 99.3 & 99.5 & 99.8 & 99.1 \\
            (69,89)  & 97.4 & 99.4 & 99.7 & 99.6 & 99.0 \\
            (69,99)  & 97.2 & 99.5 & 99.6 & 99.5 & 99.0 \\
            (69,109) & 97.7 & 99.6 & 99.6 & 99.7 & 99.2 \\
            (79,89)  & 97.7 & 99.5 & 99.7 & 99.6 & 99.1 \\
            (79,99)  & 97.3 & 99.6 & 99.6 & 99.5 & 99.0 \\
            (79,109) & 97.5 & 99.6 & 99.7 & 99.6 & 99.1 \\
            (89,99)  & 97.3 & 99.6 & 99.6 & 99.6 & 99.0 \\
            (89,109) & 97.4 & 99.6 & 99.7 & 99.7 & 99.1 \\
            (99,109) & 97.4 & 99.7 & 99.7 & 99.7 & 99.1 \\
            \bottomrule
        \end{tabular}

    \end{minipage}%
    \caption{Accuracy (\%) of the proposed texture representation $\Bold{\Upsilon}_{\omega_1, \omega_2}(\Bold{I})$ for all the benchmark datasets. The left table presents the accuracies for the random embedding sizes combinations where $\omega_1 \in \{9, 19, 29\}$. The right table exhibits the accuracies for the remaining of the combinations of the previously defined space for $(\omega_1, \omega_2)$.}
    \label{table:combined_parameter_analysis}
\end{table*}

Following this, among the RNN-based approaches, the main ones to be compared, include: SSR \cite{ribas2024color}, which combines complex network theory with the randomized neural network, the SST \cite{fares2024new} that performs a spatio-spectral texture learning by cross-channel predicting, and the VCTex \cite{fares2025volumetric} that uses volumetric (3D) color cubes for simultaneous color-texture encoding, thereby requiring only a single neural network for learning the raw texture representation, differing from the last two approaches that require more.

\begin{SCtable*}[][t]
    \centering

    \begin{tabular}{c c c c c c c}
        \toprule
        Method & Outex13 & CUReT & USPtex & MBT  & $\overline{\text{Acc}}$ & \# \\ 
        \hline

        Opponent-Gabor$^{\dagger}$ & 93.5 & 95.8 & 99.1 & 97.6 & 96.5 & $-$ \\

        LPQ$^{\dagger,i}$ & 80.1 & 91.7 & 90.4 & 95.7 & 89.5 & $-$ \\

        CLBP$^{\dagger,i}$ & 89.6 & 91.8 & 97.4 & 98.2 & 94.3 & $-$ \\
        
        CNTD$^{\dagger,i}$ & 92.3  & 91.9  & 97.9   & 98.5 & 95.2 & $-$ \\
        
        MCND$^\dagger$ & 95.4  & 97.0  & 99.0   & 97.1 & 97.1 & $-$ \\

        SSN & 96.8 & 98.6 & 99.5 & 99.0 & 98.5 & 648 \\ 

        SSR & 96.7 & 98.6 & 99.3 & 98.2 & 98.2 & 630 \\

        SST & 96.8 & 95.6$^*$ & 99.1 & 99.8 & 97.8 & 756 \\

        VCTex & 96.0 & 96.0$^*$ & 99.6 & 99.1 & 97.7 & 648 \\

        \rowcolor{gray!10} \textbf{Mixer (Ours)} & \textbf{97.8} & \textbf{99.5} & \textbf{99.7} & \underline{99.7} & \textbf{99.2} & 600 \\
        
        \bottomrule
    \end{tabular}

    \caption{Accuracy comparison with literature methods. Approaches with a $\dagger$ symbol were sourced from \cite{scabini2020spatio}, and those with symbol $^i$ indicate an integrative approach. The results with the symbol $*$ were calculated. The remaining approaches' results were obtained from their original papers. $\overline{\text{Acc}}$ refers to the average accuracy among the benchmark datasets.}
    \label{table:literature_comparison}
\end{SCtable*}

Table \ref{table:literature_comparison} presents the accuracy comparison among the literature methods. In Outex, our novel approach attained the highest accuracy of 97.8\%, being 1.0\% greater than SSN, the second-best technique, and being the unique approach to surpass the 97\% barrier, and almost the 98\%. Similarly to Outex, in CUReT, our approach also achieved the highest accuracy of 99.5\%, being 0.9\% greater than the two runner-up approaches, which were SSN and SSR, achieving 98.6\%. Thus, in both discussed datasets, the better performances indicate increases of approximately 13, and 50 images correctly classified, respectively, with all this achieved using a smaller number of descriptors.

In USPtex and MBT, our proposed technique achieved the highest accuracy on the former and the second-best performance on the latter.
The increase and the respective decrease were minimal, a difference of only 0.1\% in both datasets. 
Thus, taking into account implementation and hardware variability, it is reasonable to state that no significant improvements or declines were observed. 
However, it should be noted that our approach achieved the best accuracy in the last benchmarks with a smaller number of features, indicating that it provides a more efficient descriptor.

Finally, our approach presented the highest average accuracy of 99.2\% among the compared methods, representing an increase of 0.7\% over the second-best accuracy of 98.5\% achieved by SSN, and being the only technique to surpass the 99\%. Therefore, this emphasizes its greater discriminative power and its adaptability to diverse textural patterns and conditions present in the tackled benchmarks.

\section{Conclusions and Future Works} \label{sect:conclusions_and_future_works}

In this work, we introduced \textsc{Mixer}, a texture representation learning approach that learns by decoding the encoded texture information from hyperspherical random embeddings. By leveraging two learning branches, called \textsc{Direct} and \textsc{Mixed} branches, trained by distinct optimization objectives, and using as representation the weights of the learned linear decoder networks, our approach was able to learn a robust and efficient texture descriptor, achieving the highest average accuracy in the texture benchmarks.

To achieve this, we first densely extract local patches from the image to capture the raw texture information. Following, these patches are randomly encoded and restricted to lie on the surface of a unit hypersphere, emphasizing the directional information. Next, the \textsc{Direct} module learns to reconstruct the original patch information just by looking at its own representation. Conversely, the \textsc{Mixed} module learns to reconstruct the original patch of a specific channel by mixing the internal representation of the same patch in all image channels, thus performing a simultaneous cross-channel reconstruction by means of an intermediate fused representation.

\specialparagraph{Future Works} There are two interesting points for future works. First, we believe that there might be better ways to create the input and output matrices by leveraging distinct texture modeling techniques, such as those using \emph{complex networks}. Further, modeling these matrices differently imposes the network to cross this information, which might produce better representations. Second, we suggest that directly adjusting the optimization problem accounting for a multi-objective least squares, maintaining the closed-form expressiveness, might produce interesting representations depending on how each objective is specified.

\section*{Acknowledgements}

R. T. Fares acknowledges support from FAPESP (grant \#2024/01744-8), L. C. Ribas acknowledges support from FAPESP (grants \#2023/04583-2 and 2018/22214-6).

\bibliography{example_paper}

\begin{thebibliography}{50}
\providecommand{\natexlab}[1]{#1}
\providecommand{\url}[1]{\texttt{#1}}
\expandafter\ifx\csname urlstyle\endcsname\relax
  \providecommand{\doi}[1]{doi: #1}\else
  \providecommand{\doi}{doi: \begingroup \urlstyle{rm}\Url}\fi

\bibitem[Abdelmounaime \& Dong-Chen(2013)Abdelmounaime and Dong-Chen]{abdelmounaime2013new}
Abdelmounaime, S. and Dong-Chen, H.
\newblock New brodatz-based image databases for grayscale color and multiband texture analysis.
\newblock \emph{International Scholarly Research Notices}, 2013\penalty0 (1):\penalty0 876386, 2013.

\bibitem[Akiva et~al.(2022)Akiva, Purri, and Leotta]{akiva2022self}
Akiva, P., Purri, M., and Leotta, M.
\newblock Self-supervised material and texture representation learning for remote sensing tasks.
\newblock In \emph{Proceedings of the IEEE/CVF Conference on Computer Vision and Pattern Recognition}, pp.\  8203--8215, 2022.

\bibitem[Backes et~al.(2012)Backes, Casanova, and Bruno]{backes2012color}
Backes, A.~R., Casanova, D., and Bruno, O.~M.
\newblock Color texture analysis based on fractal descriptors.
\newblock \emph{Pattern Recognition}, 45\penalty0 (5):\penalty0 1984--1992, 2012.

\bibitem[Backes et~al.(2013)Backes, Casanova, and Bruno]{backes2013texture}
Backes, A.~R., Casanova, D., and Bruno, O.~M.
\newblock Texture analysis and classification: A complex network-based approach.
\newblock \emph{Information Sciences}, 219:\penalty0 168--180, 2013.

\bibitem[Bovik et~al.(1990)Bovik, Clark, and Geisler]{bovik1990multichannel}
Bovik, A., Clark, M., and Geisler, W.
\newblock Multichannel texture analysis using localized spatial filters.
\newblock \emph{IEEE Transactions on Pattern Analysis and Machine Intelligence}, 12\penalty0 (1):\penalty0 55--73, 1990.

\bibitem[Calvetti et~al.(2000)Calvetti, Morigi, Reichel, and Sgallari]{calvetti2000tikhonov}
Calvetti, D., Morigi, S., Reichel, L., and Sgallari, F.
\newblock Tikhonov regularization and the l-curve for large discrete ill-posed problems.
\newblock \emph{Journal of Computational and Applied Mathematics}, 123\penalty0 (1-2):\penalty0 423--446, 2000.

\bibitem[Chen et~al.(2021)Chen, Li, Quan, Xu, and Ji]{chen2021deep}
Chen, Z., Li, F., Quan, Y., Xu, Y., and Ji, H.
\newblock Deep texture recognition via exploiting cross-layer statistical self-similarity.
\newblock In \emph{Proceedings of the IEEE/CVF Conference on Computer Vision and Pattern Recognition}, pp.\  5231--5240, 2021.

\bibitem[Chen et~al.(2024)Chen, Quan, Xu, Jin, and Xu]{chen2024enhancing}
Chen, Z., Quan, Y., Xu, R., Jin, L., and Xu, Y.
\newblock Enhancing texture representation with deep tracing pattern encoding.
\newblock \emph{Pattern Recognition}, 146:\penalty0 109959, 2024.

\bibitem[Cimpoi et~al.(2015)Cimpoi, Maji, and Vedaldi]{cimpoi2015deep}
Cimpoi, M., Maji, S., and Vedaldi, A.
\newblock Deep filter banks for texture recognition and segmentation.
\newblock In \emph{Proceedings of the IEEE Conference on Computer Vision and Pattern Recognition}, pp.\  3828--3836, 2015.

\bibitem[Csurka et~al.(2004)Csurka, Dance, Fan, Willamowski, and Bray]{csurka2004visual}
Csurka, G., Dance, C., Fan, L., Willamowski, J., and Bray, C.
\newblock Visual categorization with bags of keypoints.
\newblock In \emph{Workshop on Statistical Learning in Computer Vision, ECCV}, volume~1, pp.\  1--2. Prague, 2004.

\bibitem[Dana et~al.(1999)Dana, Van~Ginneken, Nayar, and Koenderink]{dana1999reflectance}
Dana, K.~J., Van~Ginneken, B., Nayar, S.~K., and Koenderink, J.~J.
\newblock Reflectance and texture of real-world surfaces.
\newblock \emph{ACM Transactions On Graphics (TOG)}, 18\penalty0 (1):\penalty0 1--34, 1999.

\bibitem[Daugman(1985)]{daugman1985uncertainty}
Daugman, J.~G.
\newblock Uncertainty relation for resolution in space, spatial frequency, and orientation optimized by two-dimensional visual cortical filters.
\newblock \emph{Journal of the Optical Society of America A}, 2\penalty0 (7):\penalty0 1160--1169, 1985.

\bibitem[Fares \& Ribas(2024)Fares and Ribas]{fares2024new}
Fares, R.~T. and Ribas, L.~C.
\newblock A new approach to learn spatio-spectral texture representation with randomized networks: Application to brazilian plant species identification.
\newblock In \emph{International Conference on Engineering Applications of Neural Networks}, pp.\  435--449. Springer, 2024.

\bibitem[Fares \& Ribas(2025)Fares and Ribas]{fares2025volumetric}
Fares, R.~T. and Ribas, L.~C.
\newblock Volumetric color-texture representation for colorectal polyp classification in histopathology images.
\newblock In \emph{20th International Conference on Computer Vision Theory and Applications}, pp.\  210--221, 2025.

\bibitem[Florindo \& Laureano(2023)Florindo and Laureano]{florindo2023boff}
Florindo, J.~B. and Laureano, E.~E.
\newblock Boff: A bag of fuzzy deep features for texture recognition.
\newblock \emph{Expert Systems with Applications}, 219:\penalty0 119627, 2023.

\bibitem[Fujiwara \& Hashimoto(2020)Fujiwara and Hashimoto]{fujiwara2020neural}
Fujiwara, K. and Hashimoto, T.
\newblock Neural implicit embedding for point cloud analysis.
\newblock In \emph{Proceedings of the IEEE/CVF Conference on Computer Vision and Pattern Recognition}, pp.\  11734--11743, 2020.

\bibitem[Guo et~al.(2010)Guo, Zhang, and Zhang]{guo2010completed}
Guo, Z., Zhang, L., and Zhang, D.
\newblock A completed modeling of local binary pattern operator for texture classification.
\newblock \emph{IEEE Transactions on Image Processing}, 19\penalty0 (6):\penalty0 1657--1663, 2010.

\bibitem[Haralick et~al.(1973)Haralick, Shanmugam, and Dinstein]{haralick1973textural}
Haralick, R.~M., Shanmugam, K., and Dinstein, I.
\newblock Textural features for image classification.
\newblock \emph{IEEE Transactions on Systems, Man, and Cybernetics}, SMC-3\penalty0 (6):\penalty0 610--621, 1973.

\bibitem[Huang et~al.(2006)Huang, Zhu, and Siew]{huang2006extreme}
Huang, G.-B., Zhu, Q.-Y., and Siew, C.-K.
\newblock Extreme learning machine: theory and applications.
\newblock \emph{Neurocomputing}, 70\penalty0 (1-3):\penalty0 489--501, 2006.

\bibitem[Jain \& Healey(1998)Jain and Healey]{jain1998multiscale}
Jain, A. and Healey, G.
\newblock A multiscale representation including opponent color features for texture recognition.
\newblock \emph{IEEE Transactions on Image Processing}, 7\penalty0 (1):\penalty0 124--128, 1998.

\bibitem[Joanes \& Gill(1998)Joanes and Gill]{joanes1998comparing}
Joanes, D.~N. and Gill, C.~A.
\newblock Comparing measures of sample skewness and kurtosis.
\newblock \emph{Journal of the Royal Statistical Society: Series D (The Statistician)}, 47\penalty0 (1):\penalty0 183--189, 1998.

\bibitem[Jones \& Palmer(1987)Jones and Palmer]{jones1987evaluation}
Jones, J.~P. and Palmer, L.~A.
\newblock An evaluation of the two-dimensional gabor filter model of simple receptive fields in cat striate cortex.
\newblock \emph{Journal of Neurophysiology}, 58\penalty0 (6):\penalty0 1233--1258, 1987.

\bibitem[Julesz(1981)]{julesz1981textons}
Julesz, B.
\newblock Textons, the elements of texture perception, and their interactions.
\newblock \emph{Nature}, 290\penalty0 (5802):\penalty0 91--97, 1981.

\bibitem[Junior \& Backes(2016)Junior and Backes]{junior2016elm}
Junior, J. J. d. M.~S. and Backes, A.~R.
\newblock Elm based signature for texture classification.
\newblock \emph{Pattern Recognition}, 51:\penalty0 395--401, 2016.

\bibitem[Junior et~al.(2018)Junior, Backes, and Bruno]{junior2018randomized}
Junior, J. J. d. M.~S., Backes, A.~R., and Bruno, O.~M.
\newblock Randomized neural network based descriptors for shape classification.
\newblock \emph{Neurocomputing}, 312:\penalty0 201--209, 2018.

\bibitem[Leung \& Malik(2001)Leung and Malik]{leung2001representing}
Leung, T. and Malik, J.
\newblock Representing and recognizing the visual appearance of materials using three-dimensional textons.
\newblock \emph{International Journal of Computer Vision}, 43\penalty0 (1):\penalty0 29--44, 2001.

\bibitem[Liu et~al.(2019)Liu, Chen, Fieguth, Zhao, Chellappa, and Pietik{\"a}inen]{liu2019bow}
Liu, L., Chen, J., Fieguth, P., Zhao, G., Chellappa, R., and Pietik{\"a}inen, M.
\newblock From bow to cnn: Two decades of texture representation for texture classification.
\newblock \emph{International Journal of Computer Vision}, 127\penalty0 (1):\penalty0 74--109, 2019.

\bibitem[Manjunath \& Ma(1996)Manjunath and Ma]{manjunath1996texture}
Manjunath, B. and Ma, W.
\newblock Texture features for browsing and retrieval of image data.
\newblock \emph{IEEE Transactions on Pattern Analysis and Machine Intelligence}, 18\penalty0 (8):\penalty0 837--842, 1996.

\bibitem[Oiticica et~al.(2025)Oiticica, Angelim, Soares, Soares, Proen{\c{c}}a-M{\'o}dena, Bruno, and Oliveira~Jr]{oiticica2025using}
Oiticica, P.~R., Angelim, M.~K., Soares, J.~C., Soares, A.~C., Proen{\c{c}}a-M{\'o}dena, J.~L., Bruno, O.~M., and Oliveira~Jr, O.~N.
\newblock Using machine learning and optical microscopy image analysis of immunosensors made on plasmonic substrates: Application to detect the sars-cov-2 virus.
\newblock \emph{ACS Sensors}, 10\penalty0 (2):\penalty0 1407--1418, 2025.

\bibitem[Ojala et~al.(1996)Ojala, Pietik{\"a}inen, and Harwood]{ojala1996comparative}
Ojala, T., Pietik{\"a}inen, M., and Harwood, D.
\newblock A comparative study of texture measures with classification based on featured distributions.
\newblock \emph{Pattern Recognition}, 29\penalty0 (1):\penalty0 51--59, 1996.

\bibitem[Ojala et~al.(2002{\natexlab{a}})Ojala, Maenpaa, Pietikainen, Viertola, Kyllonen, and Huovinen]{ojala2002outex}
Ojala, T., Maenpaa, T., Pietikainen, M., Viertola, J., Kyllonen, J., and Huovinen, S.
\newblock Outex-new framework for empirical evaluation of texture analysis algorithms.
\newblock In \emph{International Conference on Pattern Recognition}, volume~1, pp.\  701--706. IEEE, 2002{\natexlab{a}}.

\bibitem[Ojala et~al.(2002{\natexlab{b}})Ojala, Pietikainen, and Maenpaa]{ojala2002multiresolution}
Ojala, T., Pietikainen, M., and Maenpaa, T.
\newblock Multiresolution gray-scale and rotation invariant texture classification with local binary patterns.
\newblock \emph{IEEE Transactions on Pattern Analysis and Machine Intelligence}, 24\penalty0 (7):\penalty0 971--987, 2002{\natexlab{b}}.

\bibitem[Ojansivu \& Heikkil{\"a}(2008)Ojansivu and Heikkil{\"a}]{ojansivu2008blur}
Ojansivu, V. and Heikkil{\"a}, J.
\newblock Blur insensitive texture classification using local phase quantization.
\newblock In \emph{International Conference on Image and Signal Processing}, pp.\  236--243. Springer, 2008.

\bibitem[Pao \& Takefuji(1992)Pao and Takefuji]{pao1992functional}
Pao, Y.-H. and Takefuji, Y.
\newblock Functional-link net computing: theory, system architecture, and functionalities.
\newblock \emph{Computer}, 25\penalty0 (5):\penalty0 76--79, 1992.
\newblock \doi{10.1109/2.144401}.

\bibitem[Pao et~al.(1994)Pao, Park, and Sobajic]{pao1994learning}
Pao, Y.-H., Park, G.-H., and Sobajic, D.~J.
\newblock Learning and generalization characteristics of the random vector functional-link net.
\newblock \emph{Neurocomputing}, 6\penalty0 (2):\penalty0 163--180, 1994.

\bibitem[Penrose(1955)]{generalized1955penrose}
Penrose, R.
\newblock A generalized inverse for matrices.
\newblock \emph{Mathematical Proceedings of the Cambridge Philosophical Society}, 51\penalty0 (3):\penalty0 406–413, 1955.

\bibitem[Ribas \& Bruno(2024)Ribas and Bruno]{ribas2024learning}
Ribas, L.~C. and Bruno, O.~M.
\newblock Learning a complex network representation for shape classification.
\newblock \emph{Pattern Recognition}, 154:\penalty0 110566, 2024.

\bibitem[Ribas et~al.(2020)Ribas, Junior, Scabini, and Bruno]{ribas2020fusion}
Ribas, L.~C., Junior, J. J. d. M.~S., Scabini, L.~F., and Bruno, O.~M.
\newblock Fusion of complex networks and randomized neural networks for texture analysis.
\newblock \emph{Pattern Recognition}, 103:\penalty0 107189, 2020.

\bibitem[Ribas et~al.(2024)Ribas, Scabini, Condori, and Bruno]{ribas2024color}
Ribas, L.~C., Scabini, L.~F., Condori, R.~H., and Bruno, O.~M.
\newblock Color-texture classification based on spatio-spectral complex network representations.
\newblock \emph{Physica A: Statistical Mechanics and its Applications}, 635:\penalty0 129518, 2024.

\bibitem[Scabini et~al.(2023)Scabini, Zielinski, Ribas, Gon{\c{c}}alves, De~Baets, and Bruno]{scabini2023radam}
Scabini, L., Zielinski, K.~M., Ribas, L.~C., Gon{\c{c}}alves, W.~N., De~Baets, B., and Bruno, O.~M.
\newblock Radam: Texture recognition through randomized aggregated encoding of deep activation maps.
\newblock \emph{Pattern Recognition}, 143:\penalty0 109802, 2023.

\bibitem[Scabini et~al.(2019)Scabini, Condori, Gon{\c{c}}alves, and Bruno]{scabini2019multilayer}
Scabini, L.~F., Condori, R.~H., Gon{\c{c}}alves, W.~N., and Bruno, O.~M.
\newblock Multilayer complex network descriptors for color--texture characterization.
\newblock \emph{Information Sciences}, 491:\penalty0 30--47, 2019.

\bibitem[Scabini et~al.(2020)Scabini, Ribas, and Bruno]{scabini2020spatio}
Scabini, L.~F., Ribas, L.~C., and Bruno, O.~M.
\newblock Spatio-spectral networks for color-texture analysis.
\newblock \emph{Information Sciences}, 515:\penalty0 64--79, 2020.

\bibitem[Schmidt et~al.(1992)Schmidt, Kraaijveld, Duin, et~al.]{schmidt1992feed}
Schmidt, W.~F., Kraaijveld, M.~A., Duin, R.~P., et~al.
\newblock Feed forward neural networks with random weights.
\newblock In \emph{International Conference on Pattern Recognition}, pp.\  1--4. IEEE Computer Society Press, 1992.

\bibitem[Sharan et~al.(2013)Sharan, Liu, Rosenholtz, and Adelson]{sharan2013recognizing}
Sharan, L., Liu, C., Rosenholtz, R., and Adelson, E.~H.
\newblock Recognizing materials using perceptually inspired features.
\newblock \emph{International Journal of Computer Vision}, 103\penalty0 (3):\penalty0 348--371, 2013.

\bibitem[Su et~al.(2023)Su, Zhang, Wang, Zhang, Liu, Pietik{\"a}inen, and Liu]{su2023lightweight}
Su, Z., Zhang, J., Wang, L., Zhang, H., Liu, Z., Pietik{\"a}inen, M., and Liu, L.
\newblock Lightweight pixel difference networks for efficient visual representation learning.
\newblock \emph{IEEE Transactions on Pattern Analysis and Machine Intelligence}, 45\penalty0 (12):\penalty0 14956--14974, 2023.

\bibitem[Wang \& Isola(2020)Wang and Isola]{wang2020understanding}
Wang, T. and Isola, P.
\newblock Understanding contrastive representation learning through alignment and uniformity on the hypersphere.
\newblock In \emph{International Conference on Machine Learning}, pp.\  9929--9939. PMLR, 2020.

\bibitem[Xue et~al.(2018)Xue, Zhang, and Dana]{xue2018deep}
Xue, J., Zhang, H., and Dana, K.
\newblock Deep texture manifold for ground terrain recognition.
\newblock In \emph{Proceedings of the IEEE Conference on Computer Vision and Pattern Recognition}, pp.\  558--567, 2018.

\bibitem[Zhai et~al.(2020)Zhai, Cao, Zha, Xie, and Wu]{zhai2020deep}
Zhai, W., Cao, Y., Zha, Z.-J., Xie, H., and Wu, F.
\newblock Deep structure-revealed network for texture recognition.
\newblock In \emph{Proceedings of the IEEE/CVF Conference on Computer Vision and Pattern Recognition}, pp.\  11010--11019, 2020.

\bibitem[Zhang et~al.(2017)Zhang, Xue, and Dana]{zhang2017deep}
Zhang, H., Xue, J., and Dana, K.
\newblock Deep ten: Texture encoding network.
\newblock In \emph{Proceedings of the IEEE Conference on Computer Vision and Pattern Recognition}, pp.\  708--717, 2017.

\bibitem[Zhang et~al.(2023)Zhang, Zhang, Vasudevan, and Johnson-Roberson]{zhang2023hyperspherical}
Zhang, J., Zhang, H., Vasudevan, R., and Johnson-Roberson, M.
\newblock Hyperspherical embedding for point cloud completion.
\newblock In \emph{Proceedings of the IEEE/CVF Conference on Computer Vision and Pattern Recognition}, pp.\  5323--5332, 2023.

\end{thebibliography}
\bibliographystyle{icml2025}

%
%
\clearpage
\newpage
\appendix


\icmltitlerunning{Supplementary Material for MIXER: Mixed Hyperspherical Random Embedding Neural Network for Texture Recognition}

\setcounter{equation}{0}
\renewcommand\theequation{\Alph{section}.\arabic{equation}}

\setcounter{figure}{0}
\renewcommand\thefigure{\Alph{section}.\arabic{figure}}

\section{Local Pattern Extractor Module} \label{supp:lpe}

In this section, we explore in greater depth some nuances of the proposed Local Pattern Extractor (LPE) module, whose responsibility is to extract the local intensity patterns present in each input image channel, and subsequently use this valuable information as input to the remaining of the network.

\specialparagraph{Padding} To begin with, as expressed in the main text, this process starts by centralizing $J \times J$ patches at every pixel position for a given channel of dimensions $H \times W$ of the input image. Nevertheless, as can be observed if no action were taken in relation to the channel's original dimensions, it would not be possible to center the patches next to or at the image margin.

\begin{figure}[!htbp]
    \centering
    \includegraphics[width=0.98\linewidth]{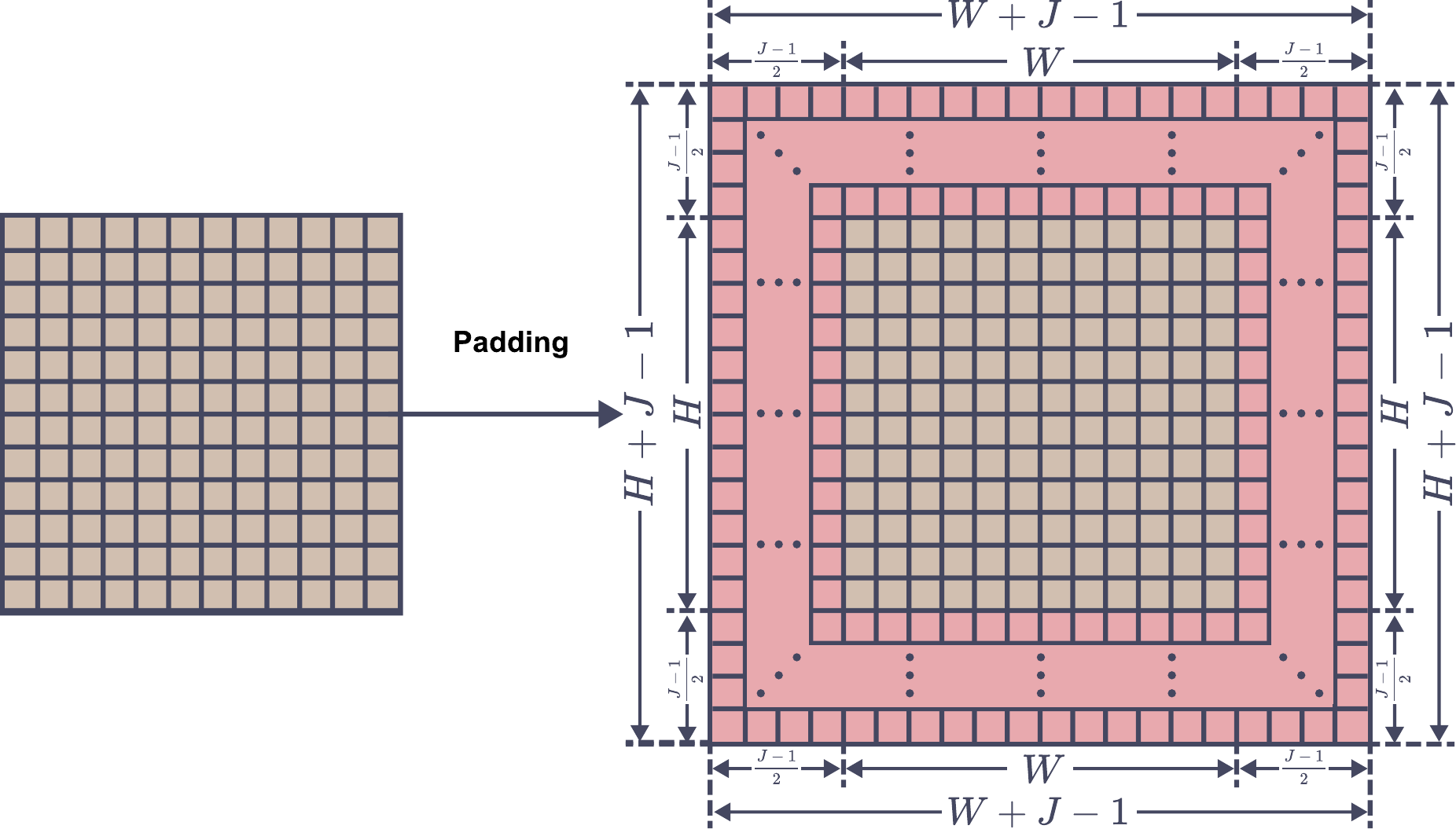}
    \caption{Illustration of the padding process of an arbitrary matrix. In this present study, this arbitrary matrix stores the pixel intensity values of an image channel.}
    \label{fig:padding_abstract_example}
\end{figure}

In this regard, to enable us to centralize the patches at every pixel position of the image channel, a padding process is performed. Thus, if the patch side is $J$, being $J$ an odd number, then a padding size of $\frac{J - 1}{2}$ is applied at every channel side. This process is depicted in Figure \ref{fig:padding_abstract_example}. As a result, after the padding process the original image channel dimensions $H \times W$ change to $(H + J - 1) \times (W + J - 1)$.

\begin{figure}[!htbp]
    \centering
    \includegraphics[width=0.98\linewidth]{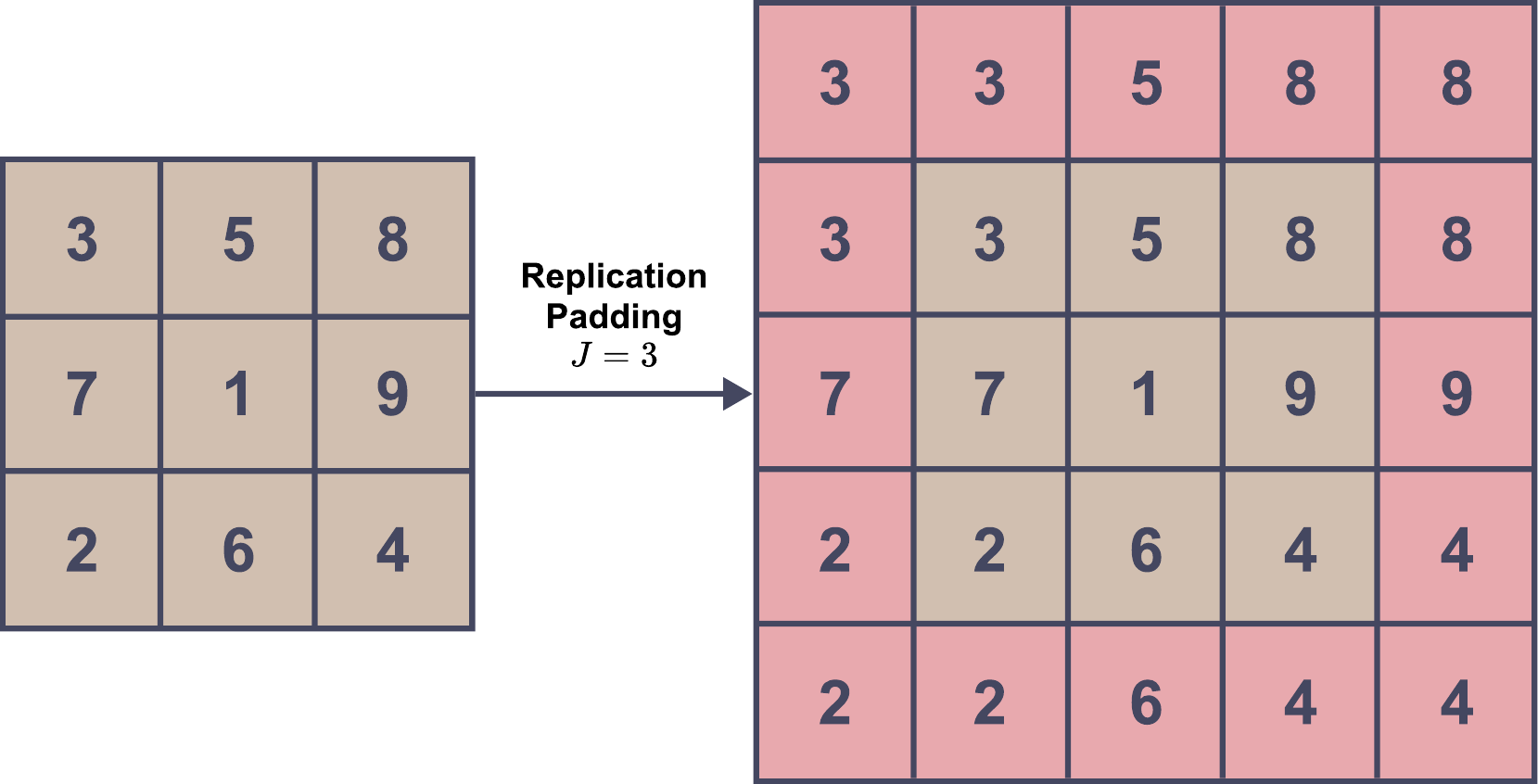}
    \caption{Concrete example of the padding process using the replication mode of a $3 \times 3$ image, and assuming a patch side size of $J = 3$. After the padding the dimension is $5 \times 5$.}
    \label{fig:padding_concrete_example}
\end{figure}

Particularly, the padding is performed employing the replication mode, where the channel's edge values are replicated. This process of replication is depicted in Figure \ref{fig:padding_concrete_example}. Furthermore, as expected, since in the concrete image example $H = W = 3$, and the patch side size is $J = 3$, the dimension after the padding process is $(H + J - 1) = (3 + 3 - 1) = 5$ (which is equal for the width dimension), matching the dimensions of the matrix of the right side of Figure \ref{fig:padding_concrete_example}.

Therefore, mathematically, the exhibited process in Figure \ref{fig:padding_concrete_example} clearly represents the input and output of the function
\begin{equation}
    \text{Pad}_J : \mathbb{R}^{C \times H \times W} \to \mathbb{R}^{C \times (H + J - 1) \times (W + J - 1)} \,.
\end{equation}
In the presented case, the input image dimension is $\Bold{I} \in \mathbb{R}^{1 \times 3 \times 3}$, and the output padded image dimension is $\text{Pad}_J(\Bold{I}) \in \mathbb{R}^{1 \times 5 \times 5}$. Thus, it should be noted that in the exemplified process the image consists of only one channel. However, for an arbitrary channel-sized image, this process is performed for each channel independently. Finally, for more information on the replication mode we refer the reader to the PyTorch v2.6.0 official documentation\footnote{https://docs.pytorch.org/docs/2.6/generated/torch.nn.functional.pad.html}.

\specialparagraph{Patch-based Extraction} After the image padding, multiple $J \times J$ patches are centered at every pixel of the padded image $\text{Pad}_J(\Bold{I})$, which corresponds to all pixels of the original image $\Bold{I}$. This is mathematically expressed by:
\begin{equation}
    \text{LPE}(\text{Pad}_J(\Bold{I})) \in \mathbb{R}^{C \times H \times W \times J \times J}\,.
\end{equation}

As expressed in the main text, this $5$-dimensional output tensor stores all of these $J \times J$ patches centered at every position for each image channel. 

\begin{figure}[t]
    \centering
    \includegraphics[width=0.98\linewidth]{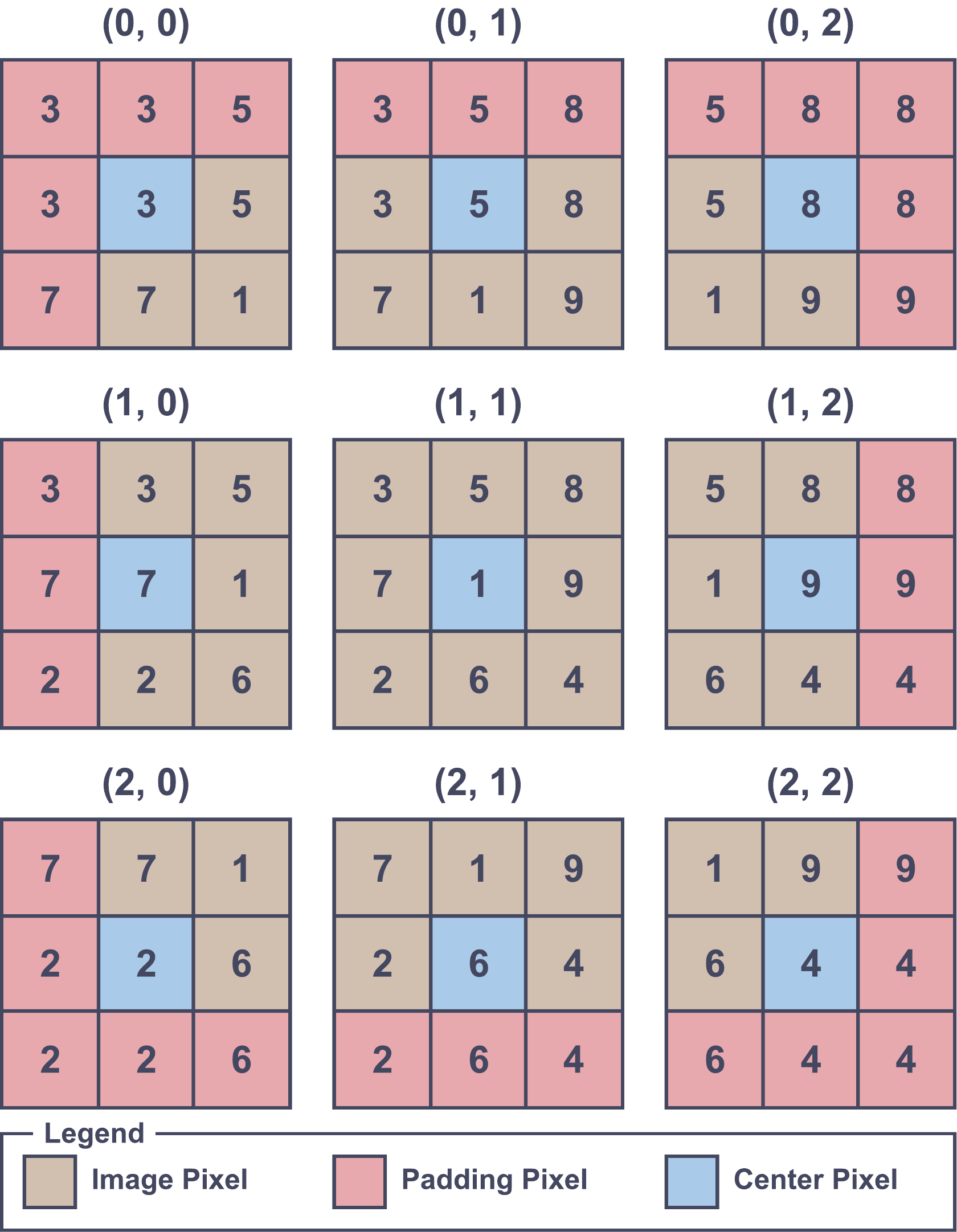}
    \caption{Concrete example of the information captured by the output of the $\textsc{LPE}$ module represented by $\text{LPE}(\text{Pad}(\Bold{I}))$. Each of the matrix represents a $3 \times 3$ patch obtained by centering it at the blue colored pixel, while also represents the content of the last two dimensions of the tensor. In addition, the overall figure can be seen as a matrix of matrices, representing the last four dimensions.}
    \label{fig:lpe_last_two_dimenions_example}
\end{figure}

In this sense, with the intent to provide a concrete example of this process. Let $\Bold{I}$ be the same image as the concrete example represented in Figure \ref{fig:padding_concrete_example}, and let $J = 3$. Hence, the Figure \ref{fig:lpe_last_two_dimenions_example} shows the all nine possible patches of size $3 \times 3$. Each of these patches represents the content of the last two dimensions of the $5$-dimensional tensor. Nevertheless, one may visualize the overall figure as a matrix of matrices, which represents the last four dimensions of that tensor.

Therefore, it can be clearly visualized how the \textsc{LPE} module captures the essential information of the raw pixels intensities information present in the input image.

\section{Hyperspherical Random Projector Module} \label{supp:rpm}

\begin{figure}[t]
    \centering
    \includegraphics[width=0.98\linewidth]{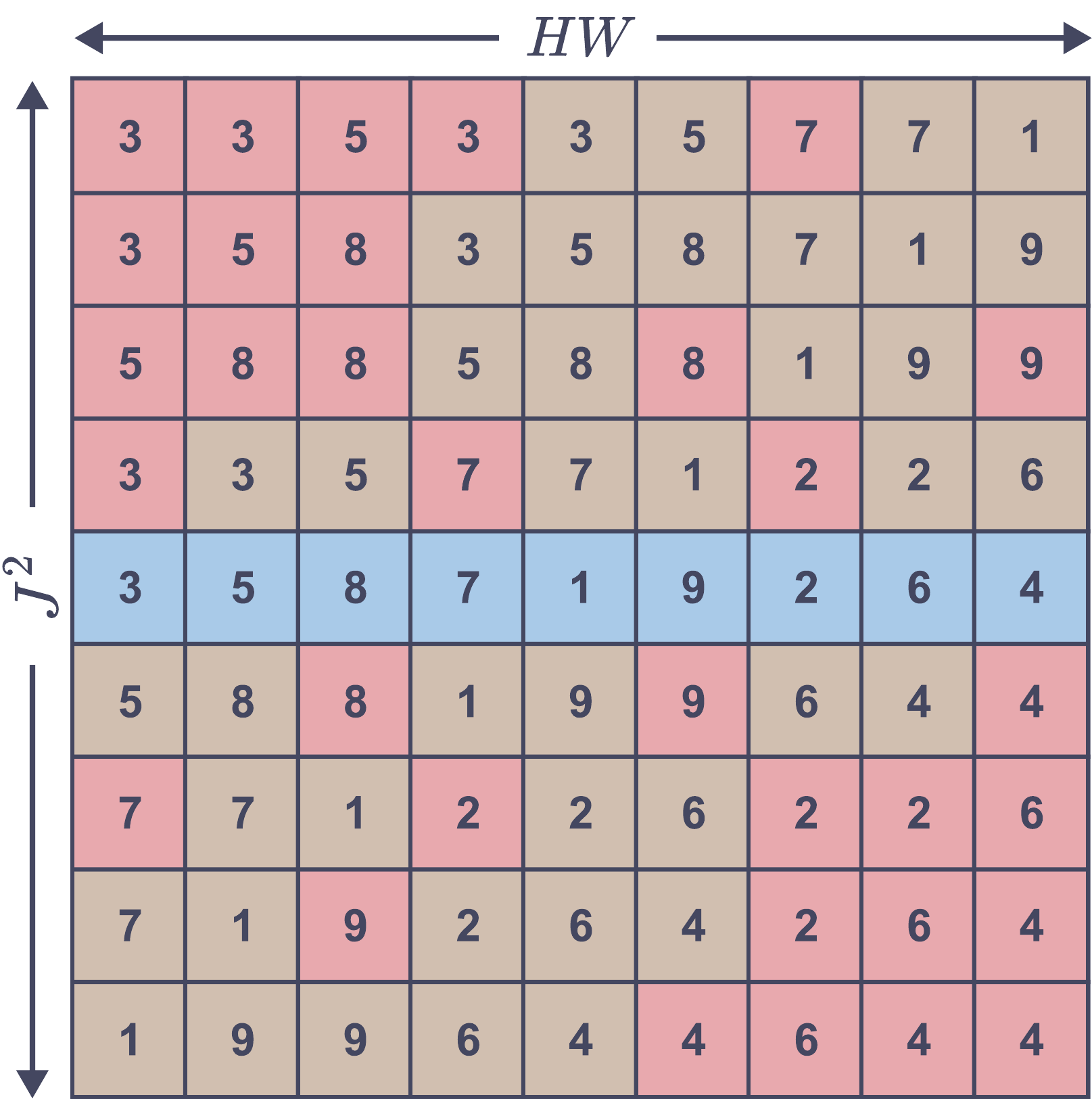}
    \caption{Concrete example of the matrix $\Bold{X} = \text{Reshape}(\Bold{L}) \in \mathbb{R}^{1 \times 9 \times 9}$ obtained after the reshaping process of the $5$-dimensional tensor $\Bold{L} \in \mathbb{R}^{1 \times 3 \times 3 \times 3 \times 3}$ represented in Figure \ref{fig:lpe_last_two_dimenions_example}.}
    \label{fig:reshaping}
\end{figure}

\specialparagraph{Reshaping} We show with a concrete example the result of the reshaping process of the output of the previous module. Let the content in Figure \ref{fig:lpe_last_two_dimenions_example} be the $5$-dimensional tensor $\Bold{L} \in \mathbb{R}^{1 \times 3 \times 3 \times 3 \times 3}$ containing the extracted $3 \times 3$ patches centered in every pixel of the image's spatial dimension. Note that the unitary dimension refers to the number of channels. For the sake of clarity, although our concrete example deals with only one channel, this is done for each channel independently. Thus, the matrix $\Bold{X} = \text{Reshape}(\Bold{L}) \in \mathbb{R}^{C \times J^2 \times HW}$ resulting from the reshaping process of the $5$-dimensional tensor $\Bold{L}$ is illustrated in Figure \ref{fig:reshaping}.

From this matrix, it can be observed that the intensities of each patch are along the column of the matrix $\Bold{X}$, while the rows refers to the intensities of a specific position inside each patch. It shall be noted that although most of the pixels are padding pixels, those colored in red, this is attributed to the fact that our initial image has a spatial dimension of $3 \times 3$. In general, for larger texture images such $128 \times 128$ most of the columns would be similar to the fifth column, i.e., without any padding pixels. To conclude, this matrix is used in the remaining of the random projector module.

\specialparagraph{Random Weight Matrix Generation} \label{supp:weight_gen} In this part, it is presented how the random weight matrix used in the process of projecting the input data into another dimensional space is built. Although in our proposed approach we need a set of these random weight matrices, i.e., one for each image channel, which would correspond to a 3-dimensional tensor, we will present the steps and an algorithm for generating a random $N$-dimensional tensor $\Bold{T} \in \mathbb{R}^{n_1 \times n_2 \times \cdots \times n_N}$.

Let $\Bold{v} \in \mathbb{R}^{L}$, where $\textstyle L = \prod_{i=1}^{N}n_i$, with $n_i$ being the size of the $i$-th dimension of $\Bold{T}$. This vector will be filled using the same procedure and configuration as in \citet{junior2016elm}. In particular, the authors used a Linear Congruential Generator (LCG) for generating pseudorandom numbers. Specifically, utilizing this approach, the elements of the vector $\Bold{v}$ satisfies the following recurrence relation:
\begin{equation}
    \Bold{v}_{k + 1} := (a \Bold{v}_k + b) \; \text{mod}\; c, \quad k \in \{0, 1, \dots, L - 2\}\,
\end{equation}
where the parameters are initialized as $\Bold{v}_0 = L + 1, a = L + 2, b = L + 3$, and $c = L^2$ following \citet{junior2016elm}. After filling the entire vector, a standardization process is applied upon $\Bold{v}$, i.e.,
\begin{equation}
    \Bold{v}_k = \frac{\Bold{v}_k - \overline{\Bold{v}}}{s_{\Bold{v}}} , \quad k \in \{0, 1, \dots, L - 1\}\,
\end{equation}
where $\textstyle \overline{\Bold{v}} = \frac{1}{L}\sum_{k=0}^{L-1}\Bold{v}_k$, is the sample mean of $\Bold{v}$, and $\textstyle s^2_{\Bold{v}} = \frac{1}{L - 1}\sum_{k=0}^{L-1}(\Bold{v}_k - \overline{\Bold{v}})^2$ is the sample variance of $\Bold{v}$.

In this context, to obtain the tensor $\Bold{T}$, the vector $\Bold{v}$ is reshaped to present $N$ dimensions, where the $i$-th dimension has $n_i$ elements. Following, we present in Algorithm \ref{supp:weight_gen_pseudocode} a PyTorch-like pseudocode for the previous taken steps.

\begin{algorithm}[tb]
    \caption{Random Weight Matrix Generator}
    \label{supp:weight_gen_pseudocode}
    \begin{algorithmic}
        \STATE {\bfseries Input:} Each dimension size $n_1, n_2, \dots, n_N$
        \STATE {\bfseries Output:} Random tensor $\Bold{T} \in \mathbb{R}^{n_1 \times n_2 \times \cdots \times n_N}$
        \STATE $\Bold{v} \leftarrow \text{torch.zeros}(\prod_{i=1}^{N}n_i)$
        \STATE $\Bold{v}_0 \leftarrow L + 1$
        \STATE $a \leftarrow L + 2$
        \STATE $b \leftarrow L + 3$
        \STATE $c \leftarrow L^2$
        \FOR {$k \leftarrow 0$ {\bfseries to} $L - 2$}
            \STATE $\Bold{v}_{k + 1} \leftarrow (a\Bold{v}_k + b) \; \text{mod} \; c$
        \ENDFOR
        \STATE $\Bold{v} \leftarrow (\Bold{v} - \text{torch.mean}(\Bold{v})) \, / \, \text{torch.std}(\Bold{v})$
        \STATE {\bfseries return} $\text{torch.reshape}(\Bold{v}, (n_1, n_2, \dots, n_N))$
    \end{algorithmic}
\end{algorithm}

In this context, if we let $\text{LCG}(n_1, n_2, \dots, n_N)$ be the $N$-dimensional tensor representing the output of the pseudocode, then for any input texture image $\Bold{I} \in \mathbb{R}^{C \times H \times W}$, the random weight matrices used in the random projector module is given by $\text{LCG}(C, \omega, J^2 + 1) \in \mathbb{R}^{C \times \omega \times (J^2 + 1)}$.



\section{Compression Module} \label{supp:comp}

In this section, we present the set $\mathcal{H}$ of compression functions used to summarize the weights of the learned decoders to assemble the texture representation. Specifically, this set consists mostly of statistical measure functions $h_1, h_2, \dots, h_{|\mathcal{H}|}$ used to compress the column vectors of the resulting matrix $\Bold{F} \in \mathbb{R}^{|\mathcal{S}|J^2 \times HW}$ from the vertical concatenation of the learned decoders' weights present in the set $\mathcal{S} = \mathcal{S}_D \cup \mathcal{S}_M$, where $\mathcal{S}_D$ and $\mathcal{S}_M$ are the sets containing the learned decoders's weights from the \textsc{Direct} and \textsc{Mixed} branches, respectively.

\begin{figure}[!htbp]
    \centering
    \includegraphics[width=0.98\linewidth]{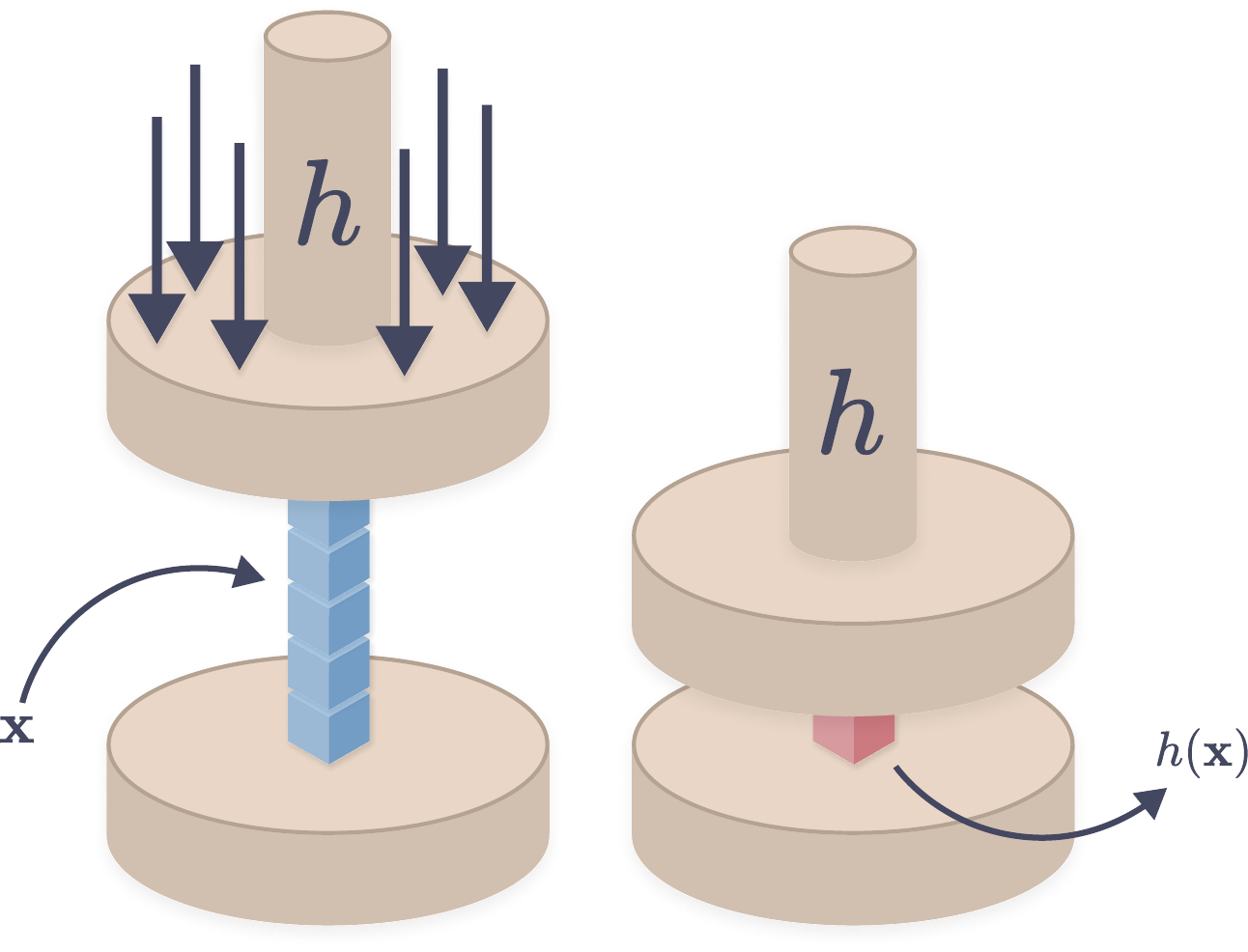}
    \caption{Illustration of a compression function $h : A \subset \mathbb{R}^{n} \to \mathbb{R}$. The compression function is responsible for compressing (or summarizing) a $n$-dimensional real vector $\Bold{x} \in A$ in a real number $h(\Bold{x})$. The complete definition is presented in the main text.}
    \label{fig:compression_function}
\end{figure}

In this context, as presented in the main text, four statistical measures are employed: mean, standard deviation, skewness, and excess kurtosis. The mean gives a central tendency of the sample, while the standard deviation is a dispersion measure. Following this, the skewness and kurtosis are measures about the shape of the distribution. In this sense, these measures complement each other, and are used to create a more robust texture representation. Next, their formulas are presented.

Let $m_r(\Bold{x})$ be the $n$-th sample central moment, which following \cite{joanes1998comparing}, is defined by:
\begin{equation}
    m_r(\Bold{x}) := \dfrac{1}{N}\sum_{k=1}^{N}\left(x_k - \overline{\Bold{x}}\right)^r \,,
\end{equation}
where $\textstyle \overline{\Bold{x}} = \frac{1}{N}\sum_{k=1}^{N}x_k$ is the sample mean of the $N$-dimensional real vector $\Bold{x}$. Thus, the utilized statistical measures' formulas are given by:
\begin{align}
    \label{eq:mean}
    h_\mu(\Bold{x}) & := \overline{\Bold{x}} \\
    \label{eq:std}
    h_\sigma(\Bold{x}) & := (m_2(\Bold{x}))^{1/2} \\
    \label{eq:skewness}
    h_\gamma(\Bold{x}) & := \dfrac{m_3(\Bold{x})}{(m_2(\Bold{x}))^{3/2}} \\
    \label{eq:kurtosis}
    h_\kappa(\Bold{x}) & := \dfrac{m_4(\Bold{x})}{(m_2(\Bold{x}))^{2}} - 3
\end{align}
It can be noted that the utilized statistical measures formulas of standard deviation and kurtosis are not unbiased estimators. Nevertheless, it is not necessary to have the most accurate estimator, the objective here is solely a mechanism to compress (summarize) an information present in a real vector to a real number.

\end{document}